\documentclass[sn-mathphys]{sn-jnl}

\usepackage{graphicx}%
\usepackage{graphicx}%
\usepackage{multirow}%
\usepackage{amsmath,amssymb,amsfonts}%
\usepackage{amsthm}%
\usepackage{mathrsfs}%
\usepackage{xcolor}%
\usepackage{textcomp}%
\usepackage{manyfoot}%
\usepackage{booktabs}%
\usepackage{algorithm}%
\usepackage{algorithmic}%
\usepackage{listings}%
\usepackage{bbding}

\theoremstyle{thmstyleone}%
\theoremstyle{thmstyletwo}%

\theoremstyle{thmstylethree}%
\begin{document}

\title[Article Title]{A Holistically Point-guided Text Framework for Weakly-Supervised Camouflaged Object Detection}




\author[1]{Tsui Qin Mok}\email{20212010144@fudan.edu.cn}
\equalcont{These authors contributed equally to this work.}

\author*[1]{Shuyong Gao}\email{sy\_gao@fudan.edu.cn}
\equalcont{These authors contributed equally to this work.}

\author[1]{Haozhe Xing}\email{hzxing21@fudan.edu.cn}

\author[1]{Miaoyang He}\email{miaoyanghe19@fudan.edu.cn}

\author[2]{Yan Wang}\email{yanwang19@fudan.edu.cn}

\author*[1,2]{Wenqiang Zhang}\email{wqzhang@fudan.edu.cn}

\affil*[1]{Shanghai Key Laboratory of Intelligent Information Processing, School of Computer Science, Fudan University, Shanghai, 200433, China}

\affil[2]{Academy for Engineering \& Technology, and the Yiwu Research Institute of Fudan University, Chengbei Road, Yiwu City, Zhejiang, 322000, China}



\abstract{Weakly-Supervised Camouflaged Object Detection (WSCOD) has gained popularity for its promise to train models with weak labels to segment objects that visually blend into their surroundings. Recently, some methods using sparsely-annotated supervision shown promising results through scribbling in WSCOD, while point-text supervision remains under-explored. Hence, this paper introduces a novel holistically point-guided text framework for WSCOD by decomposing into three phases: \textit{segment, choose, train}. Specifically, we propose Point-guided Candidate Generation (PCG), where the point’s foreground serves as a correction for the text path to explicitly correct and rejuvenate the loss detection object during the mask generation process (SEGMENT). We also introduce a Qualified Candidate Discriminator (QCD) to choose the optimal mask from a given text prompt using CLIP (CHOOSE), and employ the chosen pseudo mask for training with a self-supervised Vision Transformer (TRAIN). Additionally, we developed a new point-supervised dataset (P2C-COD) and a text-supervised dataset (T-COD). Comprehensive experiments on four benchmark datasets demonstrate our method outperforms state-of-the-art methods by a large margin, and also outperforms some existing fully-supervised camouflaged object detection methods.}

\keywords{Camouflage, Segmentation, Weakly-supervised, Multi-modal}



\maketitle

\section{Introduction}\label{sec1}

\begin{figure}[h]
    \centering
    \includegraphics[width=3.49in]{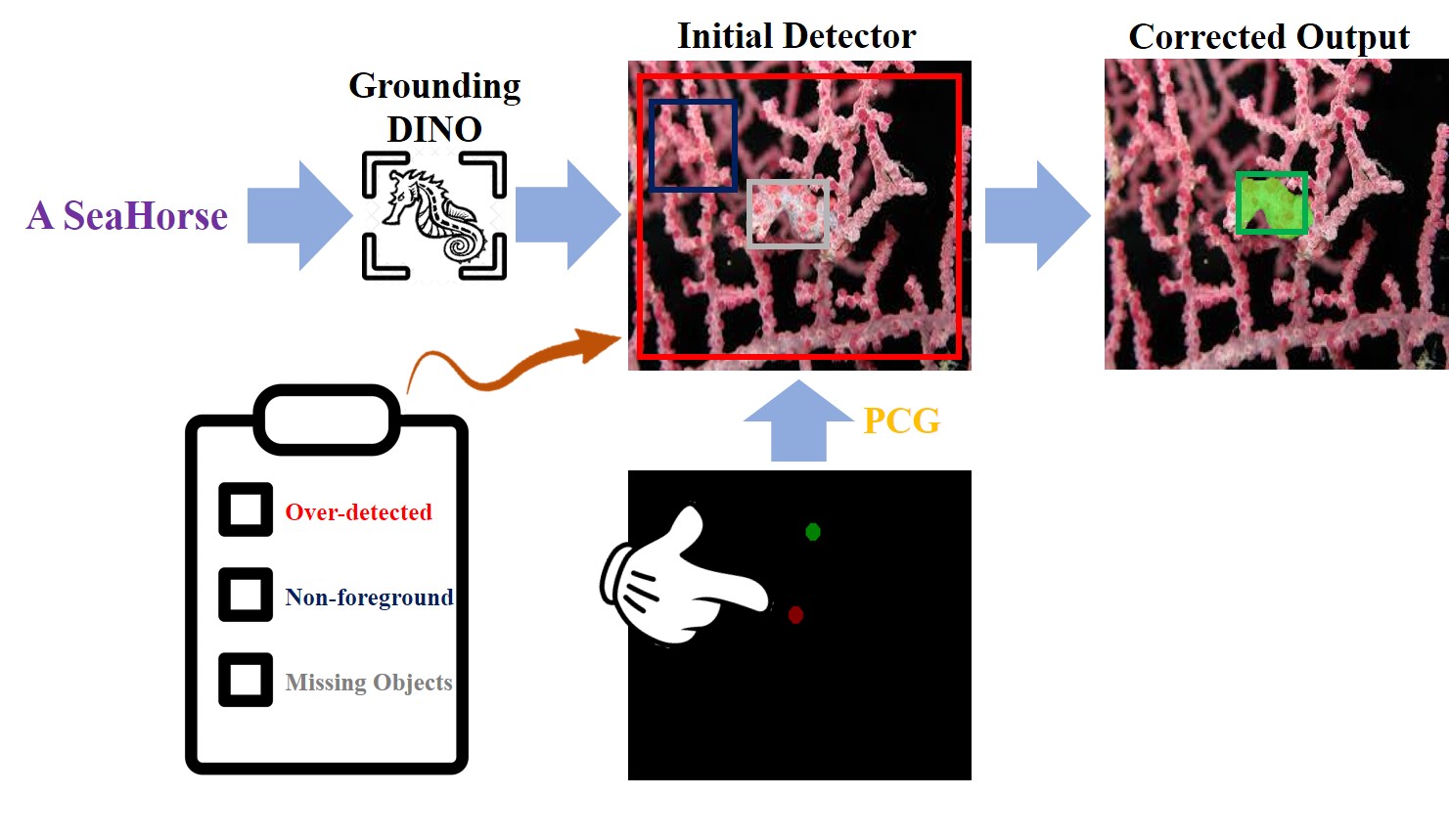}
    \caption{Illustration of our Holistically Point-guided Text framework for COD task, highlighting how the text path encounters various difficulties in identifying camouflaged objects. However, by leveraging point guidance, we can precisely correct or recover the loss detection of objects.}
    \label{figure_1}
\end{figure}

{C}{amouflaged} object detection (COD) aims to identify objects that are intentionally concealed or disguised in their surroundings \cite{FanDP2020-COD}. This task has been applied to a wide range of practical applications such as preserving wildlife, medical image segmentation (polyp segmentation \cite{FanDP2020-Pranet} or lung infection segmentation \cite{FanDP2020-InfNet}), enemy detection in the battlefield \cite{LinCJ2019-Metaheuristic}, scientific research (rare species discovery \cite{Perez2012-SpeciesDiscovery}), animal search \cite{FanDP2021-ConcealedOD}, industry (defect detection or inspection of unqualified products on the production line\cite{Zeng2022-DefectDetection}), security and surveillance systems (search-and-rescue missions, pedestrians detection or obstacles during bad weather for automatic driving), agriculture (locust detection \cite{Rustia2020-Agriculture}), and art (photo-realistic blending or recreational art \cite{Ge2018-Art}\cite{Chu2010-Art}). Camouflaged objects often exhibit complex visual patterns, texture similarities, and color matching, making them less visible and difficult to distinguish from the background \cite{Skelhorn2016-CODcognition}. Furthermore, the visibility of camouflaged objects can vary dynamically based on environmental factors, with their boundaries blending almost imperceptibly into the surrounding environment, thereby significantly complicating the detection process. Hence, rendering COD more challenging than other traditional object detection \cite{Zhao2019-TraditionalOD}. 

In recent years, there are a variety of progress in COD has been realized with the rapid development of deep convolutional neural networks (CNN) and transformers (ViT) \cite{FanDP2020-COD}\cite{FanDP2020-Pranet}\cite{FanDP2021-ConcealedOD}\cite{Sun2021-C2FNet}\cite{Lv2021-RankNet}\cite{sun2022-BGNet}\cite{Pang2022-ZoomNet}\cite{He2023-FEDER}\cite{Xing2023-SARNet}\cite{Liu2023-mscaf}, and achieved excellent performances. Nevertheless, these works heavily rely on large-scale datasets of dense pixel-level annotations, resulting in a laborious and time-consuming labeling process. To circumvent this painstaking task, this have gained significant attention from researchers to explore weakly-supervised learning to sidestep such expensive annotations by training a deep neural network with some form of abbreviated annotation that is cheaper than explicit localization cues. Some of the other related areas, such as weakly-supervised semantic segmentation \cite{Cheng2023-SemanticWSout}\cite{rong2023-SemanticWSboundary}\cite{Ru2023-SemanticWStoken}, salient object detection (SOD) \cite{veksler2023-SODWStesttimeadaptation}\cite{gao2022-SODWSpoint}\cite{cong2022-SODWSHybridLabels}\cite{yu2021-SODWSStructure}\cite{zhang2020-SODWSweaklyscribble}, visual grounding\cite{shaharabany2023-VGWSsimilarity}, remote sensing\cite{huang2022-RMSWSscribble}, and light field \cite{liang2022-LFWSweakly} have also been developed while WSCOD has not been properly explored and developed, thus remain limited progress in this field.

In the context of weakly-supervised learning, several commonly used weak labels are: 1) Bounding box supervision 2) Scribble supervision 3) Point supervision 4) Image-level labels supervision. Recently, sparse labeling methods have attracted researchers to develop methods that offer an optimal balance between time consumption and performance. Scribble annotation-based SOD, WSSA \cite{zhang2020-SODWSweaklyscribble} aims to reduce the time required for labeling. This method decreases the labeling time and offers localized ground truth labels. SCWSSOD \cite{yu2021-SODWSStructure} introduces local saliency coherence loss to take full advantage of intrinsic properties of an image that points having similar features will have similar saliency values. SCOD \cite{he2023-CODSCOD} pushes scribble work into COD to effectively detect camouflaged objects. However, scribble labeling is relatively difficult to annotate for untrained annotators where a broad range of structural information labeling could impact the efficiency and applicability of scribble annotations. PSOD \cite{gao2022-SODWSpoint} introduces the first point supervised work in SOD to solve this problem where only a single foreground and background point are annotated to an image. Nevertheless, they utilize an edge detector to obtain the structure of the objects, which is unreliable for COD due to the high intrinsic similarities between the foreground objects and background, leading to blurred edges that will decrease the final performance significantly. Despite the fact that scribble annotations provide more object information to an image, PSOD proves that point annotations are fast, and the results are promising. WSCOS \cite{he2023-CODWSCOS-SAM} leverages SAM \cite{SegmentAnything-SAM} to generate dense segmentation masks using sparse labels (e.g. scribble, points). Despite this approach's effectiveness, a large difference exists in the final results when comparing point annotations to scribble annotations.

Image-level labels are the weakest-one supervision among all other supervision methods. These weak labels do not provide any location or shape information about objects. From our research, image-level labels have not been explored in WSCOD until now. Given the context of the weakest-one and weakest sparse label supervisions, it is natural to raise a question: \textit{could we integrate these two weakest-one and weakest sparse supervisions into a unified framework to make use of each one's strengths and make them support each other?} Based on this idea, we propose a novel holistically point-guided text framework that consists of three phases, starting from the segmentation phase to generate a set of candidate masks, then the choosing phase to choose the most reliable mask, and the training phase to train our network with the final chosen pseudo mask, which is demonstrated to be effective in our experiments. 

The strength of our framework lies in its ability to generate high-quality pseudo labels for supervision by dynamically unifying predictions from two distinct paths, which significantly facilitates the accuracy of the final pseudo labels. In our work, we explore the applicability and efficacy of recent foundation models. The foundation models include SAM \cite{SegmentAnything-SAM}, Grounding DINO \cite{liu2023-GroundingDINO}, and CLIP \cite{radford2021-CLIP}, which have had a significant influence on computer vision and natural language processing. We access the potential of SAM in WSCOD by exploring two distinct settings: point input and text input in the first phase (SEGMENT). In the point input setting, we utilize our point-supervised dataset (P2C-COD) and send it directly with the RGB image into SAM to generate the first pseudo masks. Subsequently, for the text input set, we employ Grounding DINO \cite{liu2023-GroundingDINO} to generate bounding boxes of the camouflaged objects with our text-supervised dataset (T-COD) and then feed the bounding boxes and the RGB image into SAM to yield the second pseudo masks. However, since the same semantic categories may include both camouflaged and non-camouflaged objects, we find that text-generated bounding boxes from Grounding DINO could over-detect, detect non camouflaged objects, and miss any bounding boxes during the generation process due to ``camouflage" properties. To address this issue, we introduced Point-guided Candidate Generation (PCD) as shown in Figure \ref{figure_1}, in which a point's foreground acts as a guidance for text processing, facilitating the correction and elimination of flaws. Then, we conduct a re-segmentation on the corrected point-guided text path, ensuring the presence of at least one pseudo mask for the text phase. Ultimately, this results in two candidate masks, which are then advanced to the next phase, phase two (CHOOSE). Given two candidate masks generated from phase one, we choose the most appropriate one based on its alignment with the text prompt. To improve CLIP's ability to recognize objects of interest in the images, we apply a blurring effect to the background while keeping the extracted mask unaffected and in its original state in the RGB image. Subsequently, we input both images into CLIP, where their similarities to the text prompt are analyzed and compared to determine the best match. The image with the highest similarity score is then chosen as the final pseudo mask.  Finally, we train our model with the chosen pseudo mask in phase three (TRAIN). The main contributions of this work are summarized as: 
\begin{itemize}
  \item We present a novel holistically point-guided text framework for WSCOD (\textit{segment, choose, train}) to detect camouflaged objects by using point annotation and text prompt tags, and build a new point-supervised camouflaged dataset P2C-COD and a new text prompt tag camouflaged dataset T-COD.
  \item We develop a Point-guided Candidate Generation (PCG) where the point contributes the exact camouflaged object's location to text supervision to re-correct and perform re-segmentation for a new pseudo mask.
  \item We propose a simple yet efficient choosing strategy called Qualified Candidate Discriminator (QCD) to allow CLIP to recognize the objects of interest effectively, and choose the final qualified mask.
  \item Our framework greatly boosts the performance of WSCOD and surpasses the state-of-the-art methods by large margins on all six metrics for four benchmark datasets. We also outperform some existing fully-supervised camouflaged object detection methods, reducing the gap between the weakly-supervised and fully-supervised methods. 
\end{itemize}

\section{Results}\label{sec2}

\subsection{Camouflaged Object Detection}
COD is different from traditional visual detection tasks where camouflaged objects and the background have high visual discrimination, camouflaged objects tend to blend seamlessly with their surroundings, making them less distinguishable as much as possible. Therefore, the process of detecting camouflaged objects presents a more challenging task in identifying edges compared to the detection of generic or salient objects, where edges are typically more distinct and easier to discern \cite{ma2023-SODRelatedWork1} \cite{wang2023-SODRelatedWork2} \cite{liu2022-SODRelatedWork4} \cite{xie2022-SODRelatedWork5} \cite{he2023strategic}. SINet \cite{FanDP2020-COD} \cite{FanDP2021-ConcealedOD} developed a bio-inspired network to discover and locate camouflaged objects. They also proposed the largest dataset COD10K in the field of COD, which attracted many researchers to put significant efforts into COD. The emergence of this work significantly boosted the advancement of deep learning-based COD. C2F-Net \cite{Sun2021-C2FNet} enhanced COD accuracy by integrating a context-aware module that captures rich global context information. PFNet \cite{mei2021-CODPFNet} introduced the position and focus module, incorporating a distraction mining strategy to emulate human identification processes. LSR \cite{Lv2021-RankNet} presented the first multi-task framework for COD to simultaneously localize, segment, and rank camouflaged objects using a joint training strategy. SegMaR \cite{jia2022-CODSegMar} integrated a segment, magnify, and reiterate approach in a multistage, coarse-to-fine manner, effectively mimicking human behavior in interpreting complex scenarios. FEDER \cite{He2023-FEDER} proposed a network to decompose features into different frequency bands with learnable wavelets to detect camouflaged objects. In recent years, Transformer has been seen to be successful in computer vision and is known for its superior long-range modeling capabilities to capture better long-range modeling. UGTR \cite{yang2021-CODUGTR} leverages the strengths of probabilistic models and Transformer-based reasoning to learn deterministic and probabilistic information about camouflaged objects. SARNet \cite{Xing2023-SARNet} proposed a search, amplify and recognize architecture by increasing the resolution of the object area to capture camouflaged objects.

However, all these methods rely on pixel-wise annotations, which is time consuming and labor-intensive. SCOD \cite{he2023-CODSCOD} is the first weakly-supervised work in COD using scribbles by introducing two loss functions that enable the network to determine foreground and background through contrast learning and semantic analysis. WSCOS \cite{he2023-CODWSCOS-SAM} introduced a methodology that involves applying multi-augmentation views to an image and leveraging SAM \cite{SegmentAnything-SAM} to generate segmentation masks using scribble annotations. Nevertheless, we find that scribble annotations are difficult to annotate in COD, as objects blend seamlessly into the background due to their ``camouflage" characteristics. Annotators could potentially over-scribble the object along with the background. To overcome these limitations, we answer the question in Section I and proposed to utilize both the strength of point and text supervision to this work. To this extent, our approach is the first work to leverage three recent foundation models SAM \cite{SegmentAnything-SAM}, Grounding DINO \cite{liu2023-GroundingDINO}, and CLIP \cite{radford2021-CLIP} to collaborate effectively in generating and choosing the optimal pseudo mask using point annotations and text tag. We will demonstrate the superiority of our framework in facilitating these strategies in Section IV.

\subsection{Segment Anything (SAM)}
SAM \cite{SegmentAnything-SAM} is a segmentation model that was recently introduced by Meta AI Research and has gained significant attention as a large-scale model in the field of image segmentation. This model offers different prompt segmentation methods such as points, bounding boxes, and textual descriptions, and also delivers outstanding zero-shot performance to handle a wide range of scenes and objects adeptly. These two key features greatly enhance SAM's utility, making it a highly promising tool for generating various object masks in numerous computer vision applications. 

However, while it is evident that SAM performs well at segmenting images, it still faces challenges in detecting camouflaged objects \cite{tang2023can} \cite{ji2023sam}. Furthermore, its effectiveness depends on the use of carefully crafted prompts, which can be subjective and unclear. SAM-adaptor \cite{chen2023sam} overcomes these issues by utilizing a completely trained dataset of concealed objects to enhance the encoder's training. GenSAM \cite{hu2023relax} proposes a generic task prompting mechanism to generate a consensus heatmap that acts as a visual prompt to SAM to generate segmentation mask.

\subsection{Grounding DINO}
Grounding DINO \cite{liu2023-GroundingDINO} is an open-set object detector developed as an enhanced version of the transformer-based DINO detector \cite{caron2021-DINO}. Grounding DINO excels in recognizing a wide range of objects using a variety of human inputs, such as category names or referring expressions. This enhancement is enriched from its grounded pre-training, allowing it to extend its detection capabilities beyond the constraints of its training data, unlike traditional closed-set detectors, which are limited to pre-trained object categories. Furthermore, this capability enables Grounding DINO to surpass in referring object detection tasks \cite{reby2023semantic} \cite{zhang2023text2seg} \cite{chen2023weakly}, where it can interpret text categories or detailed descriptions of target objects. It detects these specified targets and delineates them by generating the minimal bounding rectangle box for each object, demonstrating a good understanding of visual and linguistic inputs. This information from Grounding DINO makes it a versatile and powerful tool in the realm of object detection and segmentation tasks.

\begin{figure}[!ht]
    \centering
    \includegraphics[width=\textwidth]{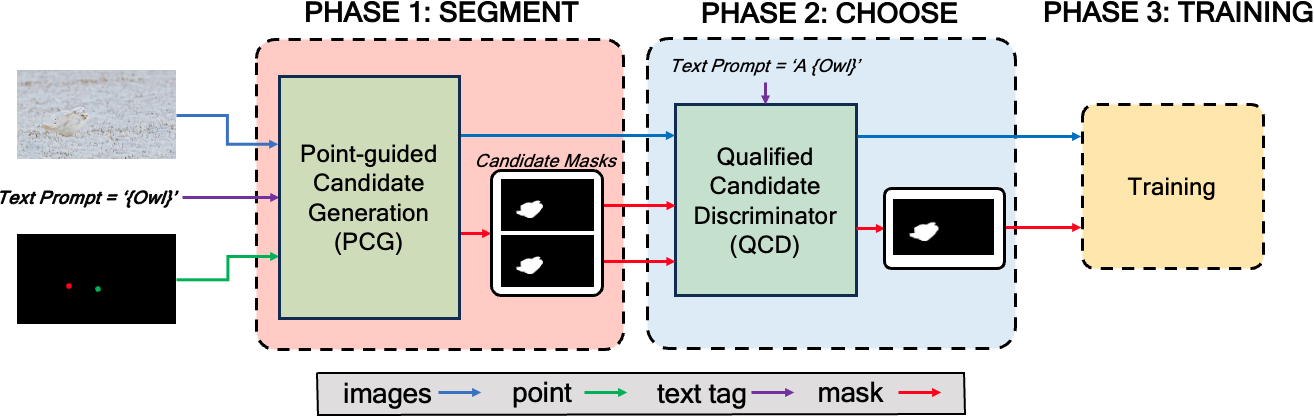}
    \caption{Our complete framework consists of three phases: \textbf{SEGMENT, CHOOSE, TRAIN} for Weakly-supervised Camouflaged Object Detection}
    \label{full_architecture}
\end{figure}

\section{Method}
Our full framework consists of three phases, which are summarized in Fig. \ref{full_architecture}. Phase 1 leverages existing foundation models to generate all segmentation masks, which we first obtain a set of two candidate masks from our developed point and text dataset within a point-guided text framework to correct bounding box errors, followed by a decision process to identify the most reliable mask among these candidates in phase 2. The final pseudo mask will be utilized as the pseudo ground-truth and input to phase 3 for training. Phases 1, 2, and 3 are described in detail in Sections A, B, and C, respectively.

\begin{figure*}[!t]
    \centering
    \includegraphics[width=\textwidth]{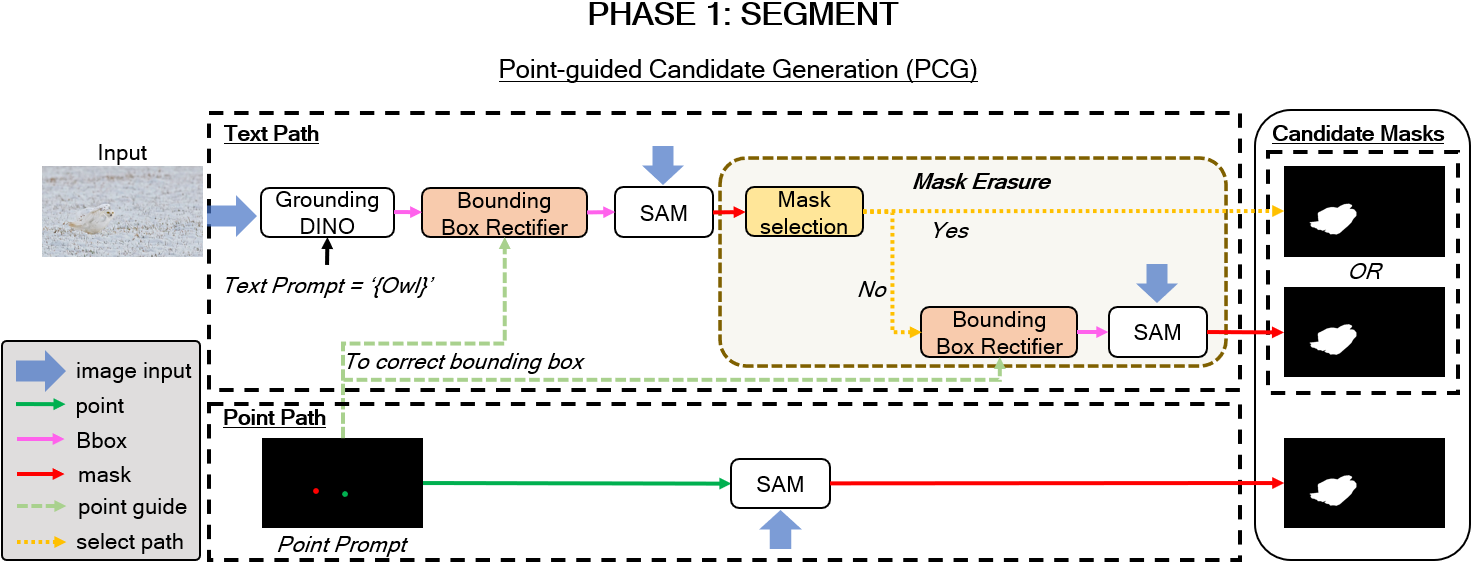}
    \caption{\textbf{\textit{Point-guided Candidate Generation (PCG)}}: Given point and text prompts, we send the prompts into the SAM \cite{SegmentAnything-SAM} to generate segmentation masks, respectively. In the text path, we first extract the bounding box, then apply point-guided bounding box correction, then determining the eligible mask through mask erasure for the object.}
    \label{phase_1_segment}
\end{figure*}

\subsection{Segment: Point-guided Candidate Generation (PCG)}
This section aims to develop a methodology for acquiring a set of segmentation masks as candidates for a given image $I \in \mathbb{R} ^{3 \times H \times W}$ using text prompt $T$ and point prompt $P(x, y)$ from our dataset, $(x, y)$ is the point's coordinates. The primary objective is to effectively combine these two points and text supervision into a unified framework, capitalizing on the unique strengths of each to create a synergistic effect where they complement and enhance each other's capabilities. Throughout this process, we assume that either one of the segmentation masks from text or point path will be the best mask for $I$. To accomplish this objective, we introduced two distinct paths (point and text) as illustrated in Fig. \ref{phase_1_segment}.

\subsubsection{Point Path}
Following the standard practices in weakly-supervised dense prediction tasks, our initial step involves generating pseudo-labels, which are then utilized for training the network. However, a challenge arises from the fact that sparse labels only encompass a limited detailed information of the object region. This constraint hinders the model’s capacity to fully capture the object's structure. Recently, a vision foundation model named SAM has been introduced, designed for the purpose of generic object segmentation \cite{SegmentAnything-SAM}, and has shown excellent results in many fields. Motivated by this, we employ a straightforward mechanism by inputting images to SAM to generate segmentation masks using our point dataset (P2C-COD) as prompt as shown in Fig. \ref{phase_1_segment}, where $M^{p} = SAM(I, P(x, y))$. In this context, $SAM(\cdot)$ represents the SAM, $M^{p}$ is the output point segmentation mask from SAM corresponding to the point path.

\subsubsection{Text Path}
After the acquisition of a candidate mask from the point path, we proceed to generate an additional candidate mask through the text path. Recently, Grounding DINO \cite{liu2023-GroundingDINO} marks a significant advancement in open-set object detection where it incorporates language to enrich its conceptual understanding, effectively merging language and vision modalities to achieve superior performance in object detection. We employ Grounding DINO for this purpose to takes in text prompt from our text dataset (T-COD) as input and returns corresponding bounding boxes for the object $B(x_{min}, y_{min}, x_{max}, y_{max}) = GD(I, T)$, where $B$ is the output bounding box for the image $I$, $T$ is the text prompt, and $GD(\cdot)$ represents the Grounding DINO, $(x_{min}, y_{min})$ are the coordinates of the top-left corner, $(x_{max}, y_{max})$ are the coordinates of the bottom-right corner of the bounding box. Considering camouflaged low-level features (texture, edge, brightness, color, and intensity features), Grounding DINO is often disrupted by these properties, leading to failures in numerous cases. To address the mistakes as shown in Fig. \ref{AS_Figure_1}, we propose a novel \textit{Bounding Box Rectifier} strategy where the core principle is to fully utilize point's strength to correct the bounding box. How to allow points to provide positive feedback to the text path is the key to success. To achieve this goal, we consider three scenarios to this strategy: 1) Generated bounding box should not occupy the entire image, as this could lead to impractically large bounding box and excessive detection of objects. This can be defined as:
\begin{equation}
\begin{aligned}
\label{PB_percentage}
PB = B \leq \alpha,
\end{aligned}
\end{equation}
\noindent where $PB$ stands for the percentage bounding box, and $\alpha$ is the proportion value of the bounding box occupied in the image. 2) Non-camouflaged objects within the image should be excluded. This is because given a text prompt (e.g. person, fish), Grounding DINO will detect both camouflaged and non-camouflaged objects, which is an outcome we seek to preclude. 3) Each of the provided point's coordinates of the camouflaged object must have at least a bounding box. Then we formulate these two scenarios as:
\begin{equation}
\begin{aligned}
\label{point_in_bbox}
x_{min} \leq x \leq x_{max}, \; y_{min} \leq y \leq y_{max},
\end{aligned}
\end{equation}
\noindent where $(x, y)$ represents all points that lie inside the bounding box, including its edges. With these three strategies, we could make use of point's advantage to regenerate the bounding box for each of the camouflaged object in the image $I$ by: 
\begin{equation}
\begin{aligned}
\label{create_bbox}
B^{rf} = x - \beta \times x, \; y - \beta \times y, \\
x + \beta \times x, \; y + \beta \times y,
\end{aligned}
\end{equation}
\noindent where $B^{rf}$ denotes the refined bounding box, $\beta$ is the scale ratio value to create the bounding box, $x$ is the horizontal coordinate of the point, and $y$ is the vertical coordinate of the point. Algorithm \ref{alg:algorithm_1} summarizes the whole process of \textit{Bounding Box Rectifier}.

\begin{algorithm}[!ht]
\caption{Bounding Box Rectifier}\label{alg:algorithm_1}
\begin{algorithmic}
\STATE \parbox[t]{0.9\linewidth} {\textbf{Input: } Image $I$, bounding box $B$, point $P$} 
\STATE \parbox[t]{0.9\linewidth} {\textbf{Output: } Refined bounding box $B^{rf}$} 
\end{algorithmic}
\begin{algorithmic}[1] 
\IF{$B$ does not satisfy condition from Eq.~\ref{PB_percentage} \textbf{or} $P$ is not within Eq.~\ref{point_in_bbox}}
    \STATE $B_{temp} \gets \text{Create new $B$ using Eq.~\ref{create_bbox} based on } I$
    \STATE $B_{rf} \gets \text{Update } B \text{ to } B_{temp}$
\ELSE
    \STATE $B_{rf} \gets B$
\ENDIF
\RETURN $B_{rf}$
\end{algorithmic}
\end{algorithm}

Once $B^{rf}$ is acquired, it is inputted into SAM to produce text segmentation masks. This process is depicted in Fig. \ref{phase_1_segment} and is represented as $M^{txt} = SAM(I, B^{rf})$, where $SAM(\cdot)$ represents the SAM, $M^{txt}$ is the resultant text segmentation mask from SAM, corresponding specifically to the text regions, $I$ symbolize the original RGB image. Although this approach allows for the generation of high-quality segmentation masks, issues of over-detection persist in objects that have yet to undergo the \textit{bounding box rectifier} correction. We then further implemented a \textit{mask erasure} technique to determine whether a mask qualifies as a final mask for its corresponding text path. We follow the same principle idea from Eq. \ref{PB_percentage} that the text segmentation mask does not cover the entire image. In cases where this occurs, we address the issue by regenerating a new bounding box using Algorithm \ref{alg:algorithm_1}. Subsequently, this bounding box is used to perform re-segmentation, resulting in a new mask. This technique is summarized in Algorithm \ref{alg:algorithm_2}.

\begin{algorithm}[!ht]
\caption{Mask Erasure}\label{alg:algorithm_2}
\begin{algorithmic}
\STATE {\textbf{Input: } Output text SAM mask $M^{txt}$, mask ratio value $\delta$}
\STATE {\textbf{Output: } Final text segmentation mask $M^{txt}_{fn}$} 
\end{algorithmic}
\begin{algorithmic}[1] 
\IF{$M^{txt} \geq \delta$}
    \STATE Apply Algorithm~\ref{alg:algorithm_1} to adjust $M^{txt}$
    \STATE $M^{txt}_{fn} \gets \text{Generated updated mask}$
\ELSE
    \STATE $M^{txt}_{fn} \gets M^{txt}$
\ENDIF
\RETURN $M^{txt}_{fn}$
\end{algorithmic}
\end{algorithm}

The output of this methodology is a set of two candidate masks for each $I$. Now we assemble both generated segmentation masks from point and text path as our candidate mask as:
\begin{equation}
\begin{aligned}
\label{candidate_mask}
C = \{ M^{pt}, M^{txt}_{fn} \},
\end{aligned}
\end{equation}
\noindent where $C$ is a set of two pseudo ground-truth masks as candidates.

\begin{figure}[!t]
    \centering
    \includegraphics[width=\textwidth]{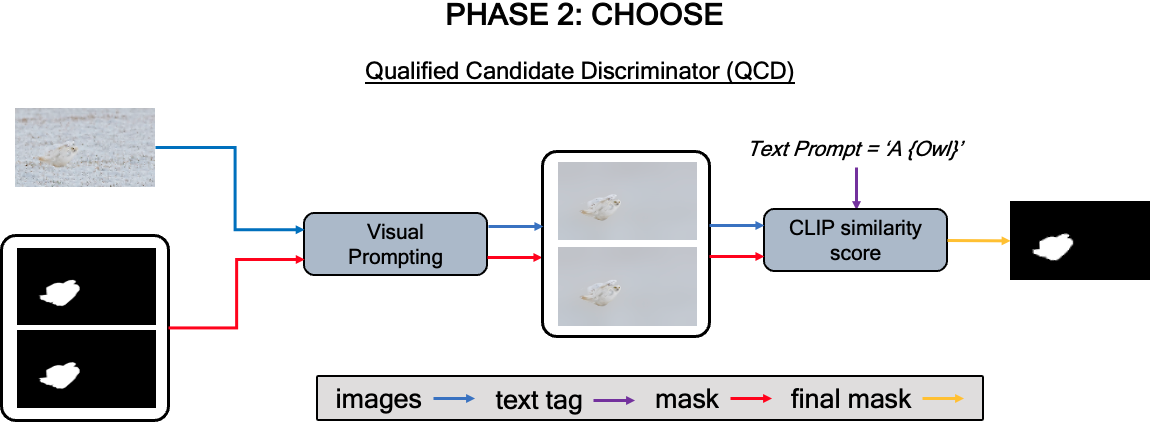}
    \caption{\textbf{\textit{Qualified Candidate Discriminator (QCD)}}: Detailed illustration of choosing the qualified mask as the final pseudo label for training.}
    \label{phase_2_choose}
\end{figure}

\subsection{Choose: Qualified Candidate Discriminator (QCD)}
Leveraging the process established in Phase 1 for image $I$, point $P$ and text $T$, we have successfully generated a set of two candidate masks, referred to as $C$. The aim of Phase 2 is to meticulously evaluate these masks to choose the one that most effectively serves as the final pseudo ground-truth for phase 3, which corresponds to the text prompt $T$ input into CLIP. We employ the image and text encoders from CLIP \cite{radford2021-CLIP}, which we refer to as:
\begin{equation}
 \phi_{\mathrm{CLIP}}: \mathbb{R}^{\mathcal{I}} \rightarrow \mathbb{R}^e \text { and } \psi_{\mathrm{CLIP}}: \mathcal{T} \rightarrow \mathbb{R}^e .
\end{equation}
\noindent Recently, \cite{yang2023-VP1} \cite{shtedritski2023-VP2} discovered an emerging ability of CLIP to link visual and textual information. The strategic use of visual prompt engineering in CLIP effectively directs the model's focus to a specific region, while simultaneously preserving the overall contextual information. \cite{shtedritski2023-VP2} propose a simple intervention method by drawing a red circle on top of the image region to extract useful emergent behaviors in CLIP. FGVP \cite{yang2023-VP1} introduce another reverse blurring visual prompting technique, where the focus is on the object instance, and the rest outside the instance mask is blurred to reduce background interference. Inspired by this, we apply the same visual prompting technique to two candidate masks on the RGB image, enabling CLIP to effectively choose the most reliable one as the final pseudo mask. Concretely, given an explicit input text prompt $\mathbf{T}^{ex}$, we compute its CLIP text embedding, $\psi_{\mathrm{CLIP}}\left(\mathbf{T}^{ex}\right)$ and pick the qualified mask that meets the following criterion: 
\begin{equation}
\max _c \operatorname{SIM}\left(\phi_{\mathrm{CLIP}}\left(\mathbf{I}^c\right), \psi_{\mathrm{CLIP}}\left(\mathbf{T}^{ex}\right)\right),
\end{equation}
\noindent where $SIM(\cdot)$ is the cosine similarity from CLIP, defined as $\operatorname{SIM}(\Tilde{u}, \Tilde{v})=\Tilde{u}^{\top} \Tilde{v} /\|\Tilde{u}\|\|\Tilde{v}\|$ for vectors $\Tilde{u}$ and $\Tilde{v}$, and $I^c$ is the visually prompted version of $I$ for $C$ using the reverse blurring technique \cite{yang2023-VP1} with a $\sigma$ = 50. Finally, we have chosen the final pseudo-label to proceed to phase 3 for training. 

\subsection{Training: Transformer-based Self-supervised DINO Model}
\begin{figure}[!t]
    \centering
    \includegraphics[width=3in]{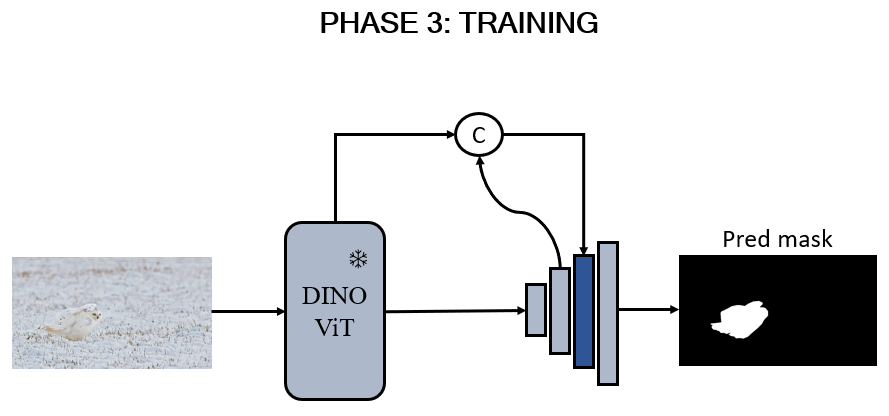}
    \caption{The framework of our training network using a self-supervised DINO transformer.}
    \label{phase_3_training}
\end{figure}

In practice, we obtain the final pseudo mask by progressing through phases 1 and 2 of our pipeline. In phase 3, as shown in Fig. \ref{phase_3_training}, we introduce a training scheme designed to train our model using the pseudo mask chosen from phase 2. Recently, Vision Transformer (ViT) \cite{dosovitskiy2020-Transformer} introduce a pure transformer model specifically for computer vision tasks, achieving unprecedented state-of-the-art performance in image classification. DINO transformer \cite{caron2021-DINO} leverages self-supervised learning to train vision transformers (ViTs) without the need for labeled data. This advancement marks a significant milestone in the field of self-supervised learning, showcasing the ability of transformers to learn powerful representations without reliance on labeled datasets. This development has motivated its application in various weakly-supervised works, as evidenced in \cite{kang2023-Distilling_w_DINO} \cite{murtaza2023-discriminative_w_DINO} \cite{gomel2023-box_w_DINO} \cite{lv2023-weaklycontrastive_w_DINO}. Inspired by this, we employ DINO ViT as our backbone to extract features, which takes an input image and produces $S = \frac{H}{8} \times \frac{W}{8}$ image (patch) tokens $P \in \mathbb{R} ^{N \times S}$ and one class token $h \in \mathbb{R} ^{N}$, where $N$ denotes the token feature dimensionality. It first transforms the image $I$ into a set of tokens, $\{P^{I}, h^{I}\}$. The multi-layer feature of the image patch tokens are denoted as $P^{I}=\{P_i | i = 1,\cdot\cdot\cdot,S^{I}\}$, where $i$ is an index over the image tokens. 

Our decoder as shown in Fig. \ref{phase_3_training}, comprises of four cascaded $3 \times 3$ convolutional layers. Each layer is sequentially followed by a Batch Normalization layer, a ReLU activation function, and an upsampling layer. This decoder architecture is designed to process the features extracted by the DINO transformer encoder. We denote the features produced at each layer of the decoder as $D=\{D_i | i = 1, 2, 3, 4\}$, where each $D_i$ corresponds to the output of the $i_{th}$ layer. Considering diverse of camouflaged properties, weak annotations often perform poorly in most cases, particularly in accurately capturing the overall structure and intricate details of objects. Hence, we merge the token containing detailed features in the shallow layer into the decoding layer as a detail supplement. Specifically, the output can be expressed as $f_e = CBR(cat(P_1, D_2))$, where $CBR$ denotes a sequence comprising a single $3 \times 3$ convolutional layer, followed by Batch Normalization and a ReLU layer. Then, we concatenate $f_e$ with $D_3$ as $cat(f_e, D_3)$ and process it through two subsequent convolution layers to yield the final multi-channel feature as $f_{f}$. Similarly to $e$, the final map $s$ can be obtained in the same manner.

\subsection{Supervision Strategy}
In our work, we employed binary cross-entropy loss (BCE) \cite{qin2019-BasNet-BCELoss}, partial binary cross-entropy loss (PBCE) \cite{tang2018-NormalizedCut-PBCELoss}, and intersection-over-union loss (IoU) \cite{mattyus2017-Deeproadmapper-IoULoss}.

The Binary Cross-Entropy (BCE) loss \cite{de2005-BCE-Introduction} is a pixel-wise loss function extensively employed in binary classification and segmentation tasks. It treats each pixel individually, ignoring the labels of neighboring pixels. Our BCE loss is defined as:
\begin{equation}
\begin{aligned}
\mathcal{L}_{BCE}= & -\sum_{i=1}^H \sum_{j=1}^WG(i, j) \log (Pr(i, j)) +(1-G(i, j)) \log (1-Pr(i, j)),
\end{aligned}
\end{equation}
\noindent \\
where $H$ and $W$ are the height and width of the images, respectively. $G(i, j)$ are the ground truth label and $Pr(i, j)$ are the predicted pixel label at the location $(i, j)$.

Partial binary cross-entropy \cite{tang2018-NormalizedCut-PBCELoss} is a specialized version of the binary cross-entropy loss, tailored for situations where only a partial of pixels in an image is labeled. Given an image with the batch of annotated pixels $B$, the partial binary cross-entropy is defined as,
\begin{equation}
\begin{aligned}
\mathcal{L}_{PBCE} =-\frac{1}{|S|} \sum_{t \in B} G(t) \log \hat{y}(t)+\left(1-G(t)\right) \log \left(1-\hat{y}(t)\right),
\end{aligned}
\end{equation}
\noindent \\
where $|S|$ is the size of the annotated pixel set, $\hat{y}(t)$ is the probability of pixel $t$ being in the target region, and $G(t)$ is the corresponding pixel ground-truth label.

Intersection over Union (IoU) initially introduced to assess the similarity between two sets \cite{jaccard1912-Distribution-IoU-Introduction}, has become a benchmark metric for evaluating performance in object detection and segmentation tasks. Its recent adaptation as a training loss function is highlighted in studies \cite{mattyus2017-Deeproadmapper-IoULoss} \cite{rahman2016-IoULoss}. To ensure its differentiability, our approach incorporates the IoU loss formulation as detailed: 
\begin{equation}
\begin{aligned}
\mathcal{L}_{IoU}=1-\frac{\sum_{i=1}^H \sum_{j=1}^W Pr(i, j) G(i, j)}{\sum_{i=1}^H \sum_{j=1}^W[Pr(i, j)+G(i, j)-Pr(i, j) G(i, j)]},
\end{aligned}
\end{equation}
\noindent \\
where $G(i, j) \in  \{0, 1\} $ is the ground-truth label of the pixel $(i, j)$ and $Pr(i, j)$ is the predicted probability of the pixel target region. $H$ and $W$ are the height and width of the input image.The overall loss of our training can be formulated as:
\begin{equation}
\begin{aligned}
\mathcal{L}_{final}= \mathcal{L}_{BCE} + \mathcal{L}_{PBCE} +\mathcal{L}_{IoU}.
\end{aligned}
\end{equation}

\section{Experiments}
In this section, we provide network architecture and implementation details for our approach and present quantitative and qualitative results to evaluate its performance experimentally. 

\subsection{Network Architecture}
In order to compare with prior works that utilized ResNet50 \cite{huang2022-RMSWSscribble} \cite{yu2021-SODWSStructure} \cite{he2023-CODSCOD} \cite{he2023-CODWSCOS-SAM} as a backbone, our approach employs a ViT-small \cite{dosovitskiy2020-Transformer} backbone comprising 21 million parameters, pretrained through self-supervision using DINO \cite{caron2021-DINO} on ImageNet 1K \cite{russakovsky2015-ImageNet}. This setup is comparable in terms of size and training data to ResNet50 \cite{he2016-ResNet}, which possesses 23 million parameters and was also trained on ImageNet 1K, but using class labels as supervision. The ViT-small backbone consists 12 transformer layers (L = 12), each with 6 attention heads (M = 6) and query-key-value embeddings of 64 dimensions (C = 64).

\subsection{Datasets}
\textbf{Point and Text-tag Supervised Datasets.}
To enhance efficiency and provide precise location data for camouflaged objects, we developed a Point-supervised Dataset (P2C-COD) and a Text-tag Supervised Dataset (T-COD) by relabeling the widely recognized CAMO \cite{le2019-Anabranch-CAMO} and COD10K \cite{FanDP2020-COD} camouflaged object detection training datasets, which comprise of 4040 training images. The annotation process for P2C-COD involved two annotators, with each image being randomly assigned one of their annotations to mitigate individual bias. In P2C-COD, only one pixel per camouflaged object is randomly chosen for labeling, and its size is enlarged for better visibility. The simplicity of this technique allows even inexperienced annotators to label an image in an average of 1 to 2 seconds. For T-COD, it focuses on identifying the types of camouflaged objects present in each image. The relabeling process for TCOD was completed in approximately one hour.

\textbf{Benchmark Datasets.}
We primarily focus on four benchmarks on our proposed approach: CAMO \cite{le2019-Anabranch-CAMO}, CHAMELEON \cite{skurowski2018-Chameleon}, COD10K \cite{FanDP2020-COD}, and NC4K \cite{Lv2021-RankNet}. CAMO dataset compromises 1,250 images, each featuring at least one camouflaged object, divided into 1,000 images for training and 250 for testing. 
CHAMELEON dataset was curated by selecting 76 images featuring camouflaged animals from Google search results. 
COD10K is currently the largest dataset in this field, containing 5,066 images of camouflaged objects, divided into 3,040 for training and 2,026 for testing, including five super-classes and 69 sub-classes. 
NC4K serves as the largest testing dataset for camouflage object segmentation, encompassing a comprehensive collection of 4,121 images, featuring both camouflage and salient objects, and art images.

\subsection{Implementation Details}
In the first phase of generating candidate masks, we employ the Grounding-DINO \cite{liu2023-GroundingDINO} and SAM (Segment Anything Model) \cite{SegmentAnything-SAM}. Specifically, we utilize the Swin-B Grounding-DINO model and ViT-H SAM model to produce segmentation masks. Subsequently, in phase two, we leverage the capabilities of CLIP (Contrastive Language–Image Pretraining) \cite{radford2021-CLIP}. Here, the ViT-L CLIP model is utilized to choose the most appropriate candidate mask from those generated in the first phase. This chosen mask is then employed as the pseudo mask for further training.

Then, we train our network using Pytorch toolbox and accelerate training with two NVIDIA GeForce RTX 3090 GPU cards. The DINO ViT-S \cite{caron2021-DINO} is in a frozen state which is used as the image encoder backbone of our network to extract encoding features, with initial parameters loaded from the pre-trained model on ImageNet \cite{russakovsky2015-ImageNet}. We set the patch size of the DINO ViT-S to 8 with a drop path rate of 0.1. The additional layers are trained using AdamW \cite{loshchilov2018-AdamW} optimizer with initial learning rate $1.0\times10^{-5}$, and follow the default PyTorch settings. Images are resized to $352\times352$ for training, and we utilized only horizontal flip for data augmentation. We set the batch size to 6 and maximum epochs to 60. Finally, we adopt poly learning rate decay for scheduling \cite{zhao2017-PolyLR} with a factor of $\left(1-\left(\frac{\text { iter }}{\text { iter }_{\max }}\right)^{0.9}\right)$ and linear warm-up for the first 12000 iterations. During testing, we resized each image to $352\times352$ and then feed it to our network to predict final output maps. 

\subsection{Evaluation Metrics}
Our methodology is evaluated using six distinct metrics: Structure-measure ($S_\alpha$) \cite{Smeasure}, Mean Absolute Error (M) \cite{MAE}, Adaptive E-measure ($E_\phi$ \cite{Emeasure} and $E_\phi^{a d}$) \cite{Emeasure2}, F-measure ($F_\phi $) \cite{Fmeasure}, maximum F-measure ($F_\beta^{\max}$), and Weighted F-measure ($F_\beta^w$) \cite{weighted-Fmeasure}. The structure-measure $S_\alpha$ is utilized to evaluate the structural integrity between the real-valued predicted map and its binary ground-truth. Mean absolute error (M) calculates the element-wise discrepancies between the predicted map and binary ground-truth. The E-measure $E_\phi$ is focused on pixel-level accuracy and overall image-level statistics. F-measure $F_\phi $ provides a comprehensive evaluation by balancing precision and recall in the prediction maps. Notably, maximum F-measure $F_\beta^{\max}$ evaluates the best balance between precision and recall achieved by the model, reflecting its peak performance. Lastly, the weighted F-measure $F_\beta^w$ adjusts the balance between precision and recall based on their relative importance in a specific context.

\subsection{Comparison With State-of-the-Art Methods}
The performance of our proposed method is compared with nine state-of-the-art (SOTA), including seven weakly-supervised methods \cite{zhang2020-SODWSweaklyscribble} \cite{piao2021-MFNet} \cite{yu2021-SODWSStructure} \cite{gao2022-SODWSpoint} \cite{he2023-CODSCOD} \cite{he2023-CODWSCOS-SAM} or two unsupervised SOD methods \cite{zhou2022-SOD-A2S} \cite{song2023-SOD-STDC}. Then, we further provide twelve fully-supervised methods for reference \cite{FanDP2020-COD} \cite{FanDP2020-Pranet} \cite{mei2021-CODPFNet} \cite{zhai2021-MGL-MutualGraphLearning} \cite{Sun2021-C2FNet} \cite{yang2021-CODUGTR} \cite{Lv2021-RankNet} \cite{zhou2022-FAPNet} \cite{zhang2022-PreyNet} \cite{jia2022-CODSegMar} \cite{Pang2022-ZoomNet} \cite{He2023-FEDER}. To ensure a fair comparison, we acquired the results of these state-of-the-art models either directly from the authors or obtained by executing the codes released by the authors. Subsequently, we evaluated them with the same evaluation tools.

\subsubsection{Quantitative Evaluation}
\begin{figure*}[htbp]
    \centering
    \vspace{0.3cm}
    \includegraphics[width=\textwidth]{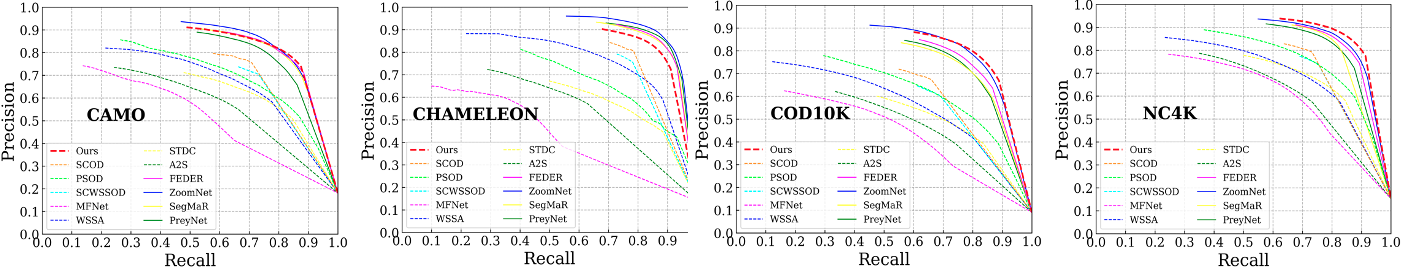}
    \caption{Illustration of precision-recall curves on four widely used COD datasets.}
    \label{pr_curves}
\end{figure*}

Table \ref{Table with others} and Figure \ref{pr_curves} present a comparative analysis with the state-of-the-art methods, where the models in the upper section are fully-supervised models, while the models in the lower section (excluding 'Ours') are weakly-supervised and unsupervised models. As shown in Table \ref{Table with others} and Figure \ref{pr_curves}, our method performs the best for six metrics on four benchmark datasets among the state-of-the-art weakly or unsupervised methods. Furthermore, our method outperforms the previous best weakly-supervised method WSCOS-S \cite{he2023-CODWSCOS-SAM} by a large margin by 3.68\% for $S_\alpha$, 1.17\% for MAE, 2.63\% for $E_\phi$, 4.15\% for $F_\phi $, 5.14\% for $F_\beta^{\max}$, and 5.97\% for $F_\beta^w$ on average for 4 compared datasets. Besides that, our method outperforms nearly all fully-supervised methods, reducing the performance gap between weakly-supervised and fully-supervised methods. A significant improvement is evident on the COD10K and NC4K (two largest testing datasets), when compared to FEDER \cite{He2023-FEDER}, which were published at CVPR 2023. On the COD10K dataset, which includes 2026 test images, our method demonstrated improvements in $S_\alpha$, $M$, $E_\phi$, $F_\phi $, $F_\beta^{\max}$ and $F_\beta^w$ metrics, with increases of 1.75\%, 0.04\%, 0.5\%, 2.25\%, 2.44\%, and 2.53\%. In contrast for NC4K, containing 4121 test images, our method exhibited enhancements in the same metrics, registering increases by 1.68\%, 0.31\%, 0.47\%, 1.98\%, 2.09\%, and 2.02\%.

\begin{table*}[htbp]
\renewcommand\arraystretch{1.6} 
\caption{Quantitative Comparison between our proposed framework and other state-of-the-art methods on four benchmarks. Top results for fully-supervised methods are shown in bold. The top two results for weakly-supervised or unsupervised methods are highlighted in {\color{red}red}, and {\color{blue}blue}.} 
\label{Table with others}
\label{compare with other sota methods}
\centering
\setlength{\tabcolsep}{1mm}
\resizebox{\linewidth}{!}{
\begin{tabular}{r|c|c|cccccc|cccccc|cccccc|cccccc}
\toprule
\multirow{2}{*}{Methods}&\multirow{2}{*}{Pub. / Year} &\multirow{2}{*}{Supervision} &\multicolumn{6}{c|}{CAMO (250 images)} &\multicolumn{6}{c|}{CHAMELEON (76 images)} &\multicolumn{6}{c|}{COD10K (2,026 images)}&\multicolumn{6}{c}{NC4K (4,121 images)}\\ 


~ & ~& ~& $S_\alpha \uparrow$ & $M \downarrow$  &$E_\phi \uparrow$  & $F_\phi \uparrow $ &$F_\beta^{max} \uparrow$ &$F_\beta^w \uparrow$
& $S_\alpha \uparrow$ & $M \downarrow$  &$E_\phi \uparrow$  & $F_\phi \uparrow $ &$F_\beta^{max} \uparrow$ &$F_\beta^w \uparrow$
& $S_\alpha \uparrow$ & $M \downarrow$  &$E_\phi \uparrow$  & $F_\phi \uparrow $ &$F_\beta^{max} \uparrow$ &$F_\beta^w \uparrow$
& $S_\alpha \uparrow$ & $M \downarrow$  &$E_\phi \uparrow$  & $F_\phi \uparrow $ &$F_\beta^{max} \uparrow$ &$F_\beta^w \uparrow$ \\ 
\midrule

SINet\cite{FanDP2020-COD} & CVPR 2020 & F & 0.7514 & 0.0996 & 0.8344 & 0.7086 & 0.7056 & 0.6055 & 0.8685 & 0.0438 & 0.8988 & 0.7755 & 0.8320 & 0.7397 & 0.7710 & 0.0511 & 0.7971 & 0.5931 & 0.6759 & 0.5509 & 0.8079 & 0.0576 & 0.8832 & 0.7681 & 0.7752 & 0.7227 \\
PraNet \cite{FanDP2020-Pranet} & MICCAI 2020 & F & 0.7798 & 0.0860 & 0.8543 & 0.7451 & 0.7486 & 0.6874 & 0.8684 & 0.0372 & 0.9174 & 0.7953 & 0.8371 & 0.7850 & 0.7942 & 0.0428 & 0.8522 & 0.6585 & 0.7144 & 0.6451 & 0.8285 & 0.0545 & 0.8863 & 0.7695 & 0.7949 & 0.7396 \\ 
PFNet \cite{mei2021-CODPFNet} & CVPR 2021 & F & 0.7822 & 0.0848 & 0.8547 & 0.7511 & 0.7582 & 0.6952 & 0.8819 & 0.0325 & 0.9418 & 0.8195 & 0.8525 & 0.8098 & 0.7998 & 0.0396 & 0.8682 & 0.6759 & 0.7247 & 0.6598 & 0.8290 & 0.0527 & 0.8937 & 0.7788 & 0.7989 & 0.7452 \\ 
MGL-R \cite{zhai2021-MGL-MutualGraphLearning} & CVPR 2021 & F & 0.7752 & 0.0883 & 0.8476 & 0.7382 & 0.7402 & 0.6733 & 0.8931 & 0.0304 & 0.9233 & 0.8256 & 0.8602 & 0.8182 & 0.8138 & 0.0351 & 0.8646 & 0.6814 & 0.7375 & 0.6725 & 0.8324 & 0.0524 & 0.8896 & 0.7782 & 0.7998 & 0.7416 \\ 
C2FNet \cite{Sun2021-C2FNet} & IJCAI 2021 & F & 0.7961 & 0.0798 & 0.8648 & 0.7642 & 0.7706 & 0.7186 & 0.8880 & 0.0316 & 0.9319 & 0.8356 & 0.8630 & 0.8284 & 0.8129 & 0.0360 & 0.8855 & 0.7027 & 0.7430 & 0.6861 & 0.8383 & 0.0490 & 0.9010 & 0.7876 & 0.8103 & 0.7623 \\
UGTR \cite{yang2021-CODUGTR} & ICCV 2021 & F & 0.7833 & 0.0863 & 0.8578 & 0.7468 & 0.7504 & 0.6834 & 0.8879 & 0.0312 & 0.9211 & 0.8045 & 0.8471 & 0.8002 & 0.8169 & 0.0355 & 0.8504 & 0.6713 & 0.7406 & 0.6728 & 0.8391 & 0.0518 & 0.8881 & 0.7782 & 0.8062 & 0.7487 \\
RankNet \cite{Lv2021-RankNet} & CVPR 2021 & F & 0.7871 & 0.0801 & 0.8591 & 0.7558 & 0.7531 & 0.6962 & 0.8898 & 0.0304 & 0.9355 & 0.8350 & 0.8619 & 0.8222 & 0.8044 & 0.0367 & 0.8828 & 0.6986 & 0.7315 & 0.6728 & 0.8395 & 0.0480 & 0.9040 & 0.8018 & 0.8146 & 0.7655 \\
FAPNet \cite{zhou2022-FAPNet} & IEEE TIP 2022 & F & \textbf{0.8150} & 0.0758 & 0.8771 & 0.7756 & 0.7918 & 0.7343 & 0.8928 & 0.0283 & 0.9252 & 0.8263 & 0.8681 & 0.8250 & 0.8221 & 0.0355 & 0.8752 & 0.7073 & 0.7577 & 0.6938 & 0.8511 & 0.0466 & 0.9029 & 0.8035 & 0.8256 & 0.7747 \\
PreyNet \cite{zhang2022-PreyNet} & ACM MM 2022 & F & 0.7895 & 0.0768 & 0.8556 & 0.7634 & 0.7652 & 0.7084 & 0.8954 & 0.0281 & 0.9512 & 0.8586 & 0.8745 & 0.8436 & 0.8129 & 0.0342 & 0.8936 & 0.7305 & 0.7474 & 0.6965 & 0.8339 & 0.0500 & 0.8965 & 0.8012 & 0.8108 & 0.7615 \\
SegMaR-1 \cite{jia2022-CODSegMar} & CVPR 2022 & F & 0.8079 & 0.0718 & 0.8703 & 0.7720 & 0.7830 & 0.7276 & 0.8915 & 0.0279 & 0.9433 & 0.8283 & 0.8654 & 0.8227 & 0.8134 & 0.0353 & 0.8808 & 0.6987 & 0.7437 & 0.6817 & 0.8405 & 0.0458 & 0.9052 & 0.8210 & 0.8264 & 0.7810 \\
ZoomNet \cite{Pang2022-ZoomNet} & CVPR 2022 & F & 0.8064 & \textbf{0.0693} & 0.8754 & \textbf{0.7934} & \textbf{0.7924} & \textbf{0.7382} & \textbf{0.9009} & \textbf{0.0238} & \textbf{0.9649} & \textbf{0.8675} & \textbf{0.8848} & \textbf{0.8503} & \textbf{0.8374} & \textbf{0.0288} & 0.8993 & \textbf{0.7522} & \textbf{0.7825} & \textbf{0.7321} & \textbf{0.8552} & \textbf{0.0412} & 0.9128 & \textbf{0.8264} & 0.8327 & \textbf{0.7910} \\ 
FEDER \cite{He2023-FEDER} & CVPR 2023 & F & 0.8021 & 0.0712 & \textbf{0.8766} & 0.7863 & 0.7886 & 0.7377 & 0.8867 & 0.0295 & 0.9429 & 0.8466 & 0.8678 & 0.8344 & 0.8223 & 0.0316 & \textbf{0.9010} & 0.7399 & 0.7678 & 0.7155 & 0.8470 & 0.0442 & \textbf{0.9129} & 0.8219 & \textbf{0.8328} & 0.7886 \\ 

\hline

A2S \cite{zhou2022-SOD-A2S} & IEEE TCSVT 2022 & U & 0.6108 & 0.1514 & 0.7332 & 0.5489 & 0.5451 & 0.4471 & 0.6433 & 0.1126 & 0.7667 & 0.5598 & 0.5586 & 0.4643 & 0.6387 & 0.0910 & 0.7142 & 0.4569 & 0.4847 & 0.4020 & 0.6768 & 0.1091 & 0.7762 & 0.6046 & 0.6134 & 0.5239 \\
STDC \cite{song2023-SOD-STDC} & ACM MM 2023 & U & 0.6582 & 0.1634 & 0.7540 & 0.6090 & 0.6007 & 0.5227 & 0.6645 & 0.1359 & 0.7344 & 0.5626 & 0.5657 & 0.4871 & 0.6548 & 0.1086 & 0.7129 & 0.4858 & 0.5132 & 0.4383 & 0.7205 & 0.1057 & 0.7960 & 0.6561 & 0.6628 & 0.5878 \\
WSSA \cite{zhang2020-SODWSweaklyscribble} & CVPR 2020 & W & 0.6876 & 0.1189 & 0.7963 & 0.6335 & 0.6338 & 0.5522 & 0.7845 & 0.0669 & 0.8592 & 0.7030 & 0.7327 & 0.6579 & 0.6910 & 0.0669 & 0.7721 & 0.5187 & 0.5492 & 0.4700 & 0.7307 & 0.0876 & 0.8206 & 0.6560 & 0.6667 & 0.5911 \\
MFNet \cite{piao2021-MFNet} & ICCV 2021 & W & 0.5728 & 0.1601 & 0.7086 & 0.5034 & 0.4892 & 0.3523 & 0.5562 & 0.1219 & 0.7235 & 0.4671 & 0.4579 & 0.3032 & 0.6219 & 0.0892 & 0.7117 & 0.4360 & 0.4470 & 0.3522 & 0.6722 & 0.1134 & 0.7668 & 0.5910 & 0.5908 & 0.4891 \\
SCWSSOD \cite{yu2021-SODWSStructure} & AAAI 2021 & W & 0.7217 & 0.1042 & 0.8258 & 0.6871 & 0.6864 & 0.6237 & 0.7957 & 0.0527 & 0.9033 & 0.7583 & 0.7644 & 0.7130 & 0.7163 & 0.0575 & 0.8263 & 0.6010 & 0.6058 & 0.5469 & 0.7625 & 0.0713 & 0.8552 & 0.7231 & 0.7248 & 0.6676 \\
PSOD \cite{gao2022-SODWSpoint} & AAAI 2022 & W & 0.6904 & 0.1249 & 0.7940 & 0.6588 & 0.6995 & 0.5332 & 0.7154 & 0.1011 & 0.7807 & 0.6058 & 0.6845 & 0.5344 & 0.7057 & 0.0682 & 0.7467 & 0.5416 & 0.6457 & 0.4844 & 0.7506 & 0.0830 & 0.8204 & 0.7057 & 0.7617 & 0.6177 \\
SCOD \cite{he2023-CODSCOD} & AAAI 2023 & W & 0.7344 & 0.0940 & 0.8337 & 0.7120 & 0.7125 & 0.6397 & 0.8136 & \textbf{\color{blue}0.0456} & 0.9008 & \textbf{\color{blue}0.7912} & \textbf{\color{blue}0.7917} & \textbf{\color{blue}0.7402} & 0.7301 & 0.0484 & 0.8404 & 0.6425 & 0.6414 & 0.5760 & 0.7735 & 0.0630 & 0.8645 & 0.7527 & 0.7525 & 0.6887 \\
WSCOS-P \cite{he2023-CODWSCOS-SAM} & NeurIPS 2023 & W & 0.7179 & 0.1019 & 0.8014 & 0.7029 & 0.6966 & 0.6020 & 0.8052 & 0.0563 & 0.8983 & 0.7673 & 0.7706 & 0.6996 & 0.7901 & 0.0387 & \textbf{\color{blue}0.8793} & \textbf{\color{blue}0.7256} & 0.7336 & 0.6634 & 0.8134 & 0.0565 & 0.8826 & 0.8007 & 0.8017 & 0.7344 \\ 
WSCOS-S \cite{he2023-CODWSCOS-SAM} & NeurIPS 2023 & W & \textbf{\color{blue}0.7585} & \textbf{\color{blue}0.0923} & \textbf{\color{blue}0.8340} & \textbf{\color{blue}0.7417} & \textbf{\color{blue}0.7373} & \textbf{\color{blue}0.6667} & \textbf{\color{blue}0.8203} & 0.0480 & \textbf{\color{blue}0.9039} & 0.7765 & 0.7753 & 0.7234 & \textbf{\color{blue}0.8026} & \textbf{\color{blue}0.0381} & 0.8704 & 0.7156 & \textbf{\color{blue}0.7400} & \textbf{\color{blue}0.6798} & \textbf{\color{blue}0.8294} & \textbf{\color{blue}0.0524} & \textbf{\color{blue}0.8910} & \textbf{\color{blue}0.8023} & \textbf{\color{blue}0.8114} & \textbf{\color{blue}0.7565} \\ 
\hline

Ours & - & W & \textbf{\color{red}0.8006} & \textbf{\color{red}0.0722} & \textbf{\color{red}0.8714} & \textbf{\color{red}0.7877} & \textbf{\color{red}0.7935} & \textbf{\color{red}0.7324} & \textbf{\color{red}0.8539} & \textbf{\color{red}0.0395} & \textbf{\color{red}0.9095} & \textbf{\color{red}0.8103} & \textbf{\color{red}0.8303} & \textbf{\color{red}0.7835} & \textbf{\color{red}0.8398} & \textbf{\color{red}0.0312} & \textbf{\color{red}0.9060} & \textbf{\color{red}0.7624} & \textbf{\color{red}0.7922} & \textbf{\color{red}0.7408} & \textbf{\color{red}0.8638} & \textbf{\color{red}0.0411} & \textbf{\color{red}0.9176} & \textbf{\color{red}0.8417} & \textbf{\color{red}0.8537} & \textbf{\color{red}0.8088} \\ 

\hline

\end{tabular}
}
\end{table*}

\subsubsection{Qualitative Evaluation}
\begin{figure*}[htbp]
\centering
\includegraphics[width=\textwidth]{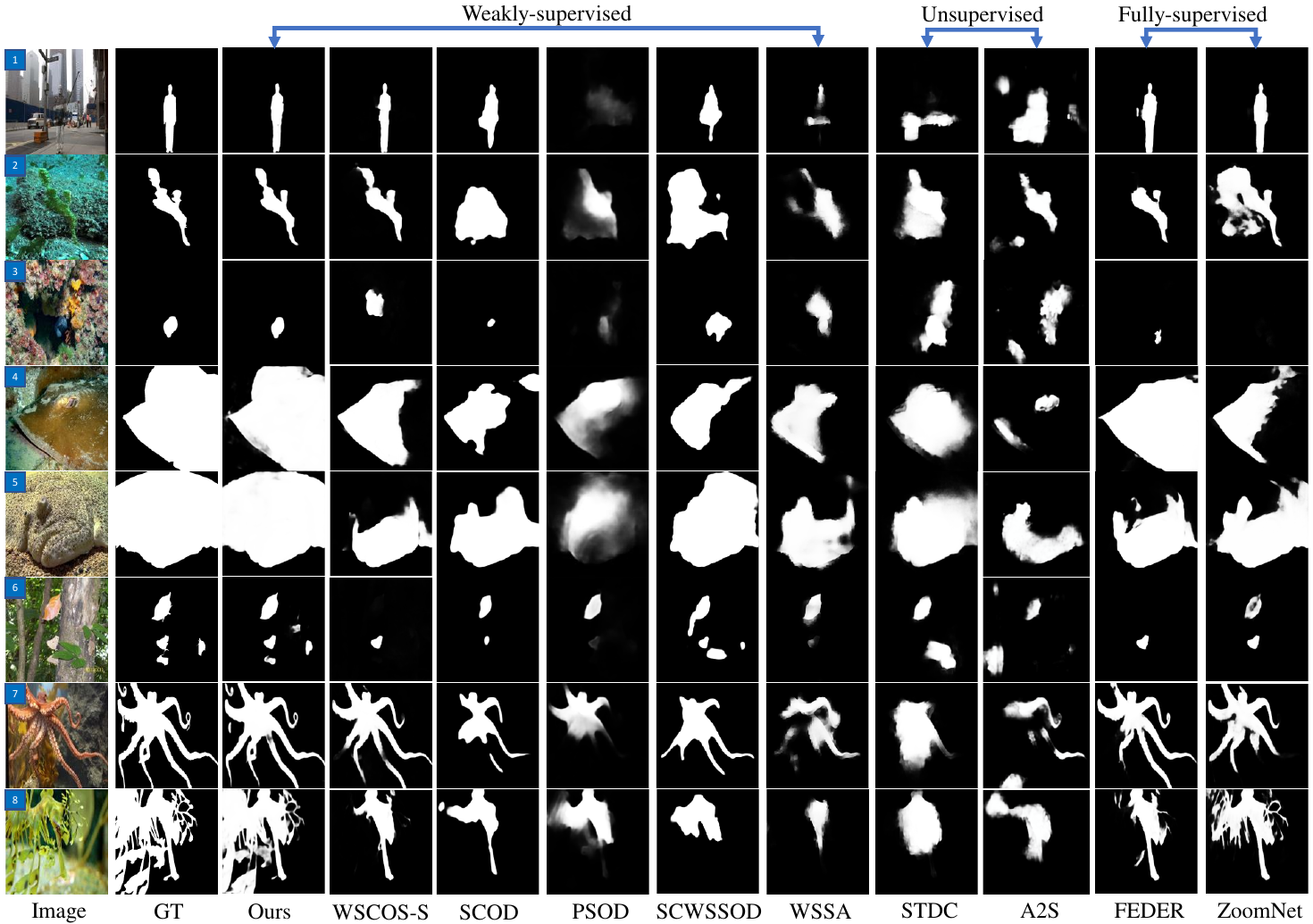}
\caption{\textbf{Visual qualitative comparison} of our method with existing state-of-the-art weakly-supervised or unsupervised methods and two fully-supervised methods. We illustrate the predictions across various complex scenarios.}
\label{visual_results}
\end{figure*}

To further demonstrate the effectiveness of our method, we present a range of qualitative results in challenging scenarios, comparing our method with existing state-of-the-art weakly-supervised or unsupervised methods, as illustrated in Fig. \ref{visual_results}. Meanwhile, we show the visualization results of two fully-supervised models for comparison. Each image is associated with different properties, including background matching (row 1), color matching (row 2), small and obscured object (row 3), large object (row 4), mimicry (row 5), multiple objects (row 6), shape complexity (row 7), and indefinable boundaries (row 8). The objective is to showcase the robustness and superior performance of our method under various conditions. It can be observed that our method accurately identifies camouflaged objects and preserves their sharp boundaries in almost all circumstances. The other methods, however, sometimes fail when dealing with complex contexts, especially when the camouflaged objects are small and obscured or dealing with multiple objects (row 3 and row 6). Thanks to our PCG where small objects can be identified accurately with clear boundaries. Furthermore, (row 7 and row 8) show that our method can accurately segment the camouflaged object on shape complexity and indefinable boundaries with fine details, and we even surpass the fully-supervised predictions from FEDER \cite{He2023-FEDER} and ZoomNet \cite{Pang2022-ZoomNet}.

\subsection{Ablation Study}
To evaluate the effectiveness of our framework, we conducted a series of ablation studies primarily on the two largest datasets (COD10K and NC4K). Only Table \ref{6th Ablation study} was evaluated using four common COD datasets. To ensure result consistency, we obtained the final performance by executing the complete pipeline comprising \textit{segment}, \textit{choose}, and \textit{training} except for Table \ref{1st Ablation study}, where the results were derived directly from phase 1. We trained the final chosen pseudo mask using a frozen DINO backbone \cite{caron2021-DINO}. To justify the design choices taken at each phase, we perform ablations on point-guided candidate generation from Phase 1, and on the qualified candidate discriminator from Phase 2.

\begin{table*}[htbp]
\caption{Ablation Study of Different Bounding Box Scale Ratio Defined in \ref{PB_percentage}.}
\label{2nd Ablation study}

\centering
\resizebox{\linewidth}{!}{
\begin{tabular}{c|cccccc|cccccc}
\toprule
\multirow{2}{*}{Scale} &\multicolumn{6}{c|}{COD10K (2,026 images)}&\multicolumn{6}{c}{NC4K (4,121 images)}\\ 


~ & $S_\alpha \uparrow$ & $M \downarrow$  &$E_\phi \uparrow$  & $F_\phi \uparrow $ &$F_\beta^{max} \uparrow$ &$F_\beta^w \uparrow$
& $S_\alpha \uparrow$ & $M \downarrow$  &$E_\phi \uparrow$  & $F_\phi \uparrow $ &$F_\beta^{max} \uparrow$ &$F_\beta^w \uparrow$ \\ 
\midrule

10 & 0.8386 & 0.0318 & 0.9023 & 0.7609 & 0.7906 & 0.7390 & 0.8644 & \textbf{0.0410} & 0.9164 & 0.8375 & 0.8523 & 0.8078 \\

\textbf{20} & \textbf{0.8398} & \textbf{0.0312} & \textbf{0.9060} & \textbf{0.7624} & \textbf{0.7922} & \textbf{0.7408} & 0.8638 & 0.0411 & \textbf{0.9176} & \textbf{0.8417} & \textbf{0.8537} & \textbf{0.8088} \\

30 & 0.8375 & 0.0320 & 0.9007 & 0.7565 & 0.7886 & 0.7363 & 0.8644 & 0.0411 & 0.9170 & 0.8365 & 0.8520 & 0.8073 \\

40 & 0.8366 & 0.0326 & 0.8978 & 0.7522 & 0.7865 & 0.7333 & 0.8657 & \textbf{0.0410} & 0.9159 & 0.8354 & 0.8533 & 0.8083 \\

50 & 0.8376 & 0.0333 & 0.8968 & 0.7525 & 0.7878 & 0.7349 & \textbf{0.8659} & \textbf{0.0410} & 0.9153 & 0.8350 & 0.8526 & 0.8083\\

\bottomrule
\end{tabular}
}
\end{table*}

\begin{figure}[t]
\centering
\includegraphics[height=3.6in, width=3.49in]{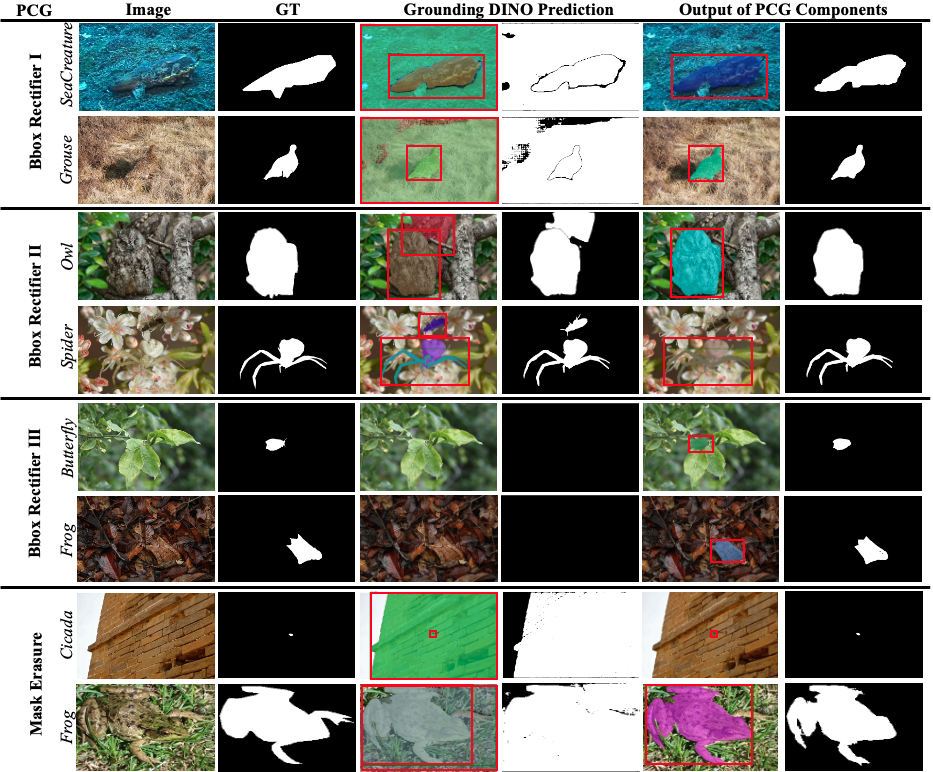}
\caption{Illustration of our \textbf{Point-guided Candidate Generation (PCG)} components to generate high quality segmentation mask. Predictions from Grounding DINO demonstrate unreliability in the fourth column, where the bounding box image and generated mask are displayed side by side. The last column also presents the bounding box image and output masks, which are produced after applying the PCG components to each respective row.}
\label{AS_Figure_1}
\end{figure}

\noindent \textbf{Effectiveness of Point-guided Candidate Generation (PCG)}. PCG encourages high segmentation masks through various corrections guided by points. To demonstrate the effectiveness of PCG, we compare it with: 1) Bounding box rectifier; 2) Mask erasure. The performance of this approach is reported in Table \ref{2nd Ablation study}. In our method, we separated the bounding box rectifier into three components: i) BR\_I; ii) BR\_II; iii) BR\_III. We did not perform ablation studies over BR\_I but set the proportion value $\alpha$ of the bounding box, as defined in Equation \ref{PB_percentage}, to be less than or equal to ninety-five percent of the total image area. This limitation is crucial because a bounding box occupying one hundred percent of the image area can cause the SAM \cite{SegmentAnything-SAM} to either over-segment detected objects or segment the entire image, neither of which aligns with our objective. We did not set the threshold for the bounding box below ninety-five percent of the total image area because camouflaged objects can be large for certain images. By maintaining this threshold, we avoid mistakenly removing true positives bounding boxes, ensuring accurate identification and segmentation of the objects. For further correction, we leave it to BR II, III and mask erasure. 

In BR\_II, we can effectively remove non-foreground objects through a straightforward, point-guided technique. If the foreground point does not fall within any bounding box, as outlined in Equation \ref{create_bbox}, we directly remove that bounding box, indicating it is a non-foreground object. However, this removal may result in no bounding boxes remaining for segmentation in some images. To address the absence of a bounding box, we leverage the strength of the point guidance to recreate a bounding box based on a scale ratio. If the scale ratio is small, it is unable to capture many large objects. In contrast, when the scale ratio is larger, it leads to the over-detection of objects, resulting in the final segmented mask containing unreasonable objects. Hence, both of these cases affect the model's performance.

\begin{table*}[t]
\renewcommand\arraystretch{1.3}
\caption{Ablation Study on Proportion Ratio of Output mask Used in Mask Erasure.}
\label{3rd Ablation study}

\centering
\setlength{\tabcolsep}{1mm}
\resizebox{\linewidth}{!}{
\begin{tabular}{c|cccccc|cccccc}
\toprule
\multirow{2}{*}{Mask Ratio} &\multicolumn{6}{c|}{COD10K (2,026 images)}&\multicolumn{6}{c}{NC4K (4,121 images)}\\ 

~ & $S_\alpha \uparrow$ & $M \downarrow$  &$E_\phi \uparrow$  & $F_\phi \uparrow $ &$F_\beta^{max} \uparrow$ &$F_\beta^w \uparrow$
& $S_\alpha \uparrow$ & $M \downarrow$  &$E_\phi \uparrow$  & $F_\phi \uparrow $ &$F_\beta^{max} \uparrow$ &$F_\beta^w \uparrow$
\\ \midrule

w/o & 0.8378 & 0.0322 & 0.9016 & 0.7574 & 0.7897 & 0.7372 & 0.8647 & 0.0412 & 0.9157 & 0.8359 & 0.8519 & 0.8072 \\

90 & 0.8374 & 0.0325 & 0.8991 & 0.7540 & 0.7886 & 0.7356 & \textbf{0.8651} & 0.0413 & 0.9157 & 0.8330 & 0.8521 & 0.8073 \\

\textbf{80} & \textbf{0.8398} & \textbf{0.0312} & \textbf{0.9060} & \textbf{0.7624} & \textbf{0.7922} & \textbf{0.7408} & 0.8638 & \textbf{0.0411} & \textbf{0.9176} & \textbf{0.8417} & \textbf{0.8537} & \textbf{0.8088} \\

70 & 0.8347 & 0.0326 & 0.8993 & 0.7561 & 0.7853 & 0.7330 & 0.8624 & 0.0417 & 0.9158 & 0.8368 & 0.8505 & 0.8057 \\

60 & 0.8358 & 0.0320 & 0.9020 & 0.7581 & 0.7878 & 0.7356 & 0.8631 & 0.0413 & 0.9163 & 0.8390 & 0.8522 & 0.8074

\\\bottomrule
\end{tabular}
}
\end{table*}

\begin{table*}[!t]
\renewcommand\arraystretch{1.3}
\caption{Ablation Study Results of PCG Components for Phase 1.}
\label{1st Ablation study}

\centering
\setlength{\tabcolsep}{1mm}
\resizebox{\linewidth}{!}{
\begin{tabular}{ccccc|cccccc|cccccc}
\toprule
\multicolumn{5}{c|}{PCG module} &\multicolumn{6}{c|}{COD10K (2,026 images)}&\multicolumn{6}{c}{NC4K (4,121 images)}\\ 


Text & BR\_I & BR\_II & BR\_III & Mask Erasure 
& $S_\alpha \uparrow$ & $M \downarrow$  &$E_\phi \uparrow$  & $F_\phi \uparrow $ &$F_\beta^{max} \uparrow$ &$F_\beta^w \uparrow$
& $S_\alpha \uparrow$ & $M \downarrow$  &$E_\phi \uparrow$  & $F_\phi \uparrow $ &$F_\beta^{max} \uparrow$ &$F_\beta^w \uparrow$
\\ \midrule

\CheckmarkBold & ~ & ~ & ~ & ~ & 0.8263 & 0.0412 & 0.8909 & 0.7327 & 0.7719 & 0.7161 & 0.8532 & 0.0518 & 0.9044 & 0.8099 & 0.8327 & 0.7867 \\

\CheckmarkBold & \CheckmarkBold & ~ & ~ & ~ & 0.8312 & 0.0338 & 0.8965 & 0.7412 & 0.7712 & 0.7213 & 0.8603 & 0.0411 & 0.9124 & 0.8206 & 0.8361 & 0.7957 \\

\CheckmarkBold & \CheckmarkBold & \CheckmarkBold & ~ & ~ & 0.8287 & 0.0331 & 0.8956 & 0.7394 & 0.7674 & 0.7186 & 0.8581 & 0.0413 & 0.9103 & 0.8190 & 0.8326 & 0.7927 \\

\CheckmarkBold & \CheckmarkBold & \CheckmarkBold & \CheckmarkBold & ~ & 0.8361 & 0.0324 & 0.9016 & 0.7518 & 0.7826 & 0.7318 & \textbf{0.8664} & 0.0399 & 0.9186 & 0.8351 & 0.8498 & 0.8081 \\

\CheckmarkBold & ~ & ~ & ~ & \CheckmarkBold & 0.8296 & 0.0360 & 0.8956 & 0.7403 & 0.7725 & 0.7216 & 0.8597 & 0.0441 & 0.9101 & 0.8187 & 0.8378 & 0.7947 \\

\CheckmarkBold & \CheckmarkBold & \CheckmarkBold & \CheckmarkBold & \CheckmarkBold & \textbf{0.8362} & \textbf{0.0322} & \textbf{0.9038} & \textbf{0.7541} & \textbf{0.7846} & \textbf{0.7338} & 0.8662 & \textbf{0.0399} & \textbf{0.9201} & \textbf{0.8356} & \textbf{0.8500} & \textbf{0.8086}\\

\bottomrule
\end{tabular}
}
\end{table*}

Some bounding boxes meet all the requirements of the bounding box rectifier and thus bypass the point-guided correction; however, the resulting output mask is still not reliable enough to serve as the final pseudo mask. We propose a novel mask erasure strategy to assess whether a mask is reliable to serve as the final pseudo mask for the corresponding text path. As demonstrated in Table \ref{3rd Ablation study}, we observe that utilizing an appropriate mask ratio better preserves true positives and eliminates true negatives, and can further alleviate mask inconsistency caused by Grounding DINO \cite{liu2023-GroundingDINO}. Since PCG comprises four components, we conducted ablation experiments to study and analyze the effectiveness of our PCG components as shown in Table \ref{1st Ablation study}. These experiments are based on the results generated in phase 1 only. The first line indicates that we directly utilize the text prompt to generate bounding box from Grounding DINO, and then use SAM to create the pseudo mask, which is subsequently employed for training. The subsequent lines display the results of our PCG components and effectively demonstrate the efficacy of our proposed PCG.

\begin{table*}[!t]
\caption{Comparison of Different Visual Engineering Prompting.}
\label{4th Ablation study}

\centering
\setlength{\tabcolsep}{1mm}
\resizebox{\linewidth}{!}{
\begin{tabular}{c|cccccc|cccccc}
\toprule
\multirow{2}{*}{Visual Prompt Choice} &\multicolumn{6}{c|}{COD10K (2,026 images)}&\multicolumn{6}{c}{NC4K (4,121 images)}\\ 

~ & $S_\alpha \uparrow$ & $M \downarrow$  &$E_\phi \uparrow$  & $F_\phi \uparrow $ &$F_\beta^{max} \uparrow$ &$F_\beta^w \uparrow$
& $S_\alpha \uparrow$ & $M \downarrow$  &$E_\phi \uparrow$  & $F_\phi \uparrow $ &$F_\beta^{max} \uparrow$ &$F_\beta^w \uparrow$
\\ \midrule

Random & 0.8171 & 0.0339 & 0.8959 & 0.7539 & 0.7694 & 0.7120 & 0.8387 & 0.0480 & 0.9017 & 0.8284 & 0.8332 & 0.7769 \\ 

Rectangle & 0.8267 & 0.0326 & 0.9031 & 0.7606 & 0.7797 & 0.7246 & 0.8406 & 0.0485 & 0.9047 & 0.8325 & 0.8358 & 0.7795 \\ 

Red Ellipse & 0.8191 & 0.0335 & 0.8999 & 0.7554 & 0.7705 & 0.7148 & 0.8429 & 0.0471 & 0.9040 & 0.8291 & 0.8337 & 0.7802 \\ 

\textbf{Reverse Blur} & \textbf{0.8398} & \textbf{0.0312} & \textbf{0.9060} & \textbf{0.7624} & \textbf{0.7922} & \textbf{0.7408} & \textbf{0.8638} & \textbf{0.0411} & \textbf{0.9176} & \textbf{0.8417} & \textbf{0.8537} & \textbf{0.8088} 

\\\bottomrule
\end{tabular}
}
\end{table*}

\begin{table*}[!t]
\renewcommand\arraystretch{1.3}
\caption{Ablation Study of Different Text Prompt to CLIP \cite{radford2021-CLIP} .}
\label{5th Ablation study}

\centering
\setlength{\tabcolsep}{1mm}
\resizebox{\linewidth}{!}{
\begin{tabular}{c|cccccc|cccccc}
\toprule
\multirow{2}{*}{Text Prompting} &\multicolumn{6}{c|}{COD10K (2,026 images)}&\multicolumn{6}{c}{NC4K (4,121 images)}\\ 

~ & $S_\alpha \uparrow$ & $M \downarrow$  &$E_\phi \uparrow$  & $F_\phi \uparrow $ &$F_\beta^{max} \uparrow$ &$F_\beta^w \uparrow$
& $S_\alpha \uparrow$ & $M \downarrow$  &$E_\phi \uparrow$  & $F_\phi \uparrow $ &$F_\beta^{max} \uparrow$ &$F_\beta^w \uparrow$
\\ \midrule

\{\textit{text}\} & 0.8386 & 0.0314 & 0.9058 & 0.7611 & 0.7903 & 0.7388 & 0.8629 & \textbf{0.0411} & 0.9177 & 0.8399 & 0.8518 & 0.8076 \\

A photo of \{\textit{text}\} & 0.8389 & \textbf{0.0312} & \textbf{0.9063} & 0.7608 & 0.7906 & 0.7387 & \textbf{0.8639} & \textbf{0.0411} & \textbf{0.9182} & 0.8410 & 0.8533 & \textbf{0.8091} \\

A \{\textit{text}\} & \textbf{0.8398} & \textbf{0.0312} & 0.9060 & \textbf{0.7624} & \textbf{0.7922} & \textbf{0.7408} & 0.8638 & \textbf{0.0411} & 0.9176 & \textbf{0.8417} & \textbf{0.8537} & 0.8088

\\\bottomrule
\end{tabular}
}
\end{table*}

\noindent \textbf{Visualization Results of PCG components}.
We visualize the output masks of different components to demonstrate the effectiveness of our PCG, as shown in Fig. \ref{AS_Figure_1}. In rows one to three, the PCG filters out unnecessary and non-foreground objects and compensates for losses or errors encountered by Grounding DINO. Even when the output mask from SAM bypasses the correction checks of the bounding box rectifier, the mask erasure performs further checks on the output mask. This is to determine whether it qualifies as the final mask for the text path, as illustrated in row four. This process showcases the PCG's ability in various scenarios, thereby providing excellent conditions for the creation of more refined segmentation masks.

\noindent \textbf{Effectiveness of Qualified Candidate Discriminator (QCD)}. In our proposed QCD module, predictions from two paths jointly participate as candidates for choosing the optimal mask. To evaluate the effectiveness of our QCD in accurately choosing the best mask among the candidate masks produced in Phase 2, we conduct ablation studies using different visual engineering prompting methods such as Reverse Blur \cite{yang2023-VP1} and Red Ellipse \cite{shtedritski2023-VP2}. Additionally, we compare these methods with simpler baselines, such as randomly selecting a mask (Random) or drawing a rectangle around the mask (Rectangle). The results of these comparisons are reported in Table \ref{5th Ablation study}. The reverse blur approach \cite{yang2023-VP1}, utilizing a CLIP ViT-L/14@336px as the visual backbone, outperforms all other visual prompting methods. This result validates our choice of approach in Phase 2. Furthermore, we conducted ablation studies on the text prompts fed into CLIP to identify camouflaged objects as shown in Table \ref{6th Ablation study}. We notice that the choice of text prompt sent to CLIP has a significant impact on overall performance. Appropriately selecting an appropriate text prompt can improve this performance, enhancing CLIP's ability to accurately recognize the objects.

\begin{figure}[!t]
    \centering
    \includegraphics[width=3.49in]{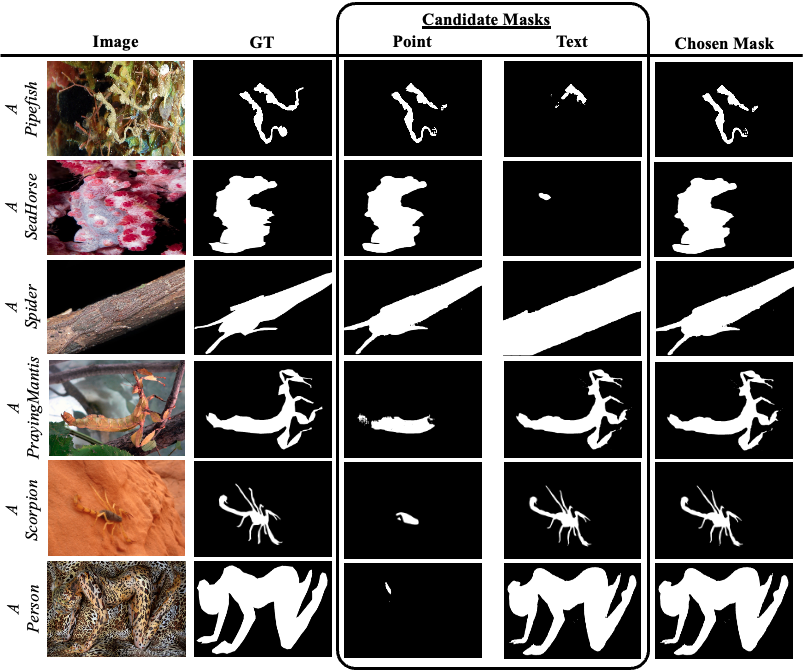}
    \caption{Illustration of the chosen masks based on the text prompt for \textbf{Qualified Candidate Discriminator (QCD)}.}
    \label{AS_Figure_2}
\end{figure}

\noindent \textbf{Visualization Results of QCD}. As shown in Fig. \ref{AS_Figure_2}, the first three rows display the point masks chosen by CLIP based on the given text prompts, while the last three rows show the text masks selected by CLIP. It is evident that when the output masks from either the point path or text path are clear, having undergone visual engineering prompting with an appropriately chosen text prompt, this enables CLIP to effectively choose the mask most similar to the specific camouflaged object. One advantage of this approach is that even if one path performs poorly in phase 1, the other path can still effectively be chosen by CLIP to produce a more accurate segmentation mask.

\begin{table*}[!t]
\renewcommand\arraystretch{1.3}
\caption{Comparisons of the quality of pseudo labels on our DINO ViT network.}
\label{7th Ablation study}

\centering
\setlength{\tabcolsep}{1mm}
\resizebox{\linewidth}{!}{
\begin{tabular}{c|cccccc|cccccc}
\toprule
\multirow{2}{*}{Methods} &\multicolumn{6}{c|}{COD10K (2,026 images)}&\multicolumn{6}{c}{NC4K (4,121 images)}\\ 

~ & $S_\alpha \uparrow$ & $M \downarrow$  &$E_\phi \uparrow$  & $F_\phi \uparrow $ &$F_\beta^{max} \uparrow$ &$F_\beta^w \uparrow$
& $S_\alpha \uparrow$ & $M \downarrow$  &$E_\phi \uparrow$  & $F_\phi \uparrow $ &$F_\beta^{max} \uparrow$ &$F_\beta^w \uparrow$
\\ \midrule

MFNet & 0.6304 & 0.0859 & 0.7281 & 0.4794 & 0.4715 & 0.3953 & 0.6850 & 0.1061 & 0.7650 & 0.6167 & 0.6122 & 0.5320 \\
PSOD & 0.7154 & 0.0590 & 0.8061 & 0.6131 & 0.6190 & 0.5499 & 0.7511 & 0.0781 & 0.8291 & 0.7348 & 0.7356 & 0.6565 \\
WSCOS-S & 0.8219 & 0.0487 & 0.8474 & 0.7096 & 0.7866 & 0.6818 & 0.8559 & 0.0561 & 0.8874 & 0.8053 & 0.8494 & 0.7648 \\ 

\hline
Ours & \textbf{0.8398} & \textbf{0.0312} & \textbf{0.9060} & \textbf{0.7624} & \textbf{0.7922} & \textbf{0.7408} & \textbf{0.8638} & \textbf{0.0411} & \textbf{0.9176} & \textbf{0.8417} & \textbf{0.8537} & \textbf{0.8088} \\

\bottomrule
\end{tabular}
}
\end{table*}

\noindent \textbf{Comparison of the quality of pseudo-labels}. In weakly supervised learning, two common approaches are used. The first generates pseudo-labels from weak annotations (e.g., image-level labels or scribbles), which are refined and used to train the network. The second trains the network directly with weak annotations, using techniques like propagation, specialized loss functions, and regularization to achieve the final performance without creating pixel-level pseudo-labels. Based on our approach, we compare methods that first generate pseudo-labels and train these pseudo-labels on our DINO ViT network. As shown in Table \ref{7th Ablation study}, we achieve the best performance for both of the largest datasets compared to the previous best WSCOS-S method, indicating that our generated pseudo-labels are of higher quality.

\noindent \textbf{Comparison of our proposed framework}. From Table \ref{6th Ablation study}, training pseudo mask generated from SAM alone using either point or text supervision yields poor results. However, with our proposed method in Phase 1, \textit{segment}, there is a noticeable improvement in all metrics. Moving on to Phase 2, there is a decline in performance on the CAMO dataset compared to Phase 1. This can be attributed to the CAMO dataset containing many large objects, blurred, and old pictures, hence leading to objects not detected by Grounding DINO or unrecognized by CLIP. Despite the implementation of our proposed bounding box rectifier, the scale ratio remains too small to effectively capture large objects. Consequently, these problems can significantly impact the final performance. 

\begin{table*}[t]
\renewcommand\arraystretch{1.3}
\caption{Ablation Studies of the Complete Point-guided Text Framework.}
\label{6th Ablation study}

\centering
\setlength{\tabcolsep}{1mm}
\resizebox{\linewidth}{!}{
\begin{tabular}{c|cccccc|cccccc|cccccc|cccccc}
\toprule
\multirow{2}{*}{Phase} &\multicolumn{6}{c|}{CAMO (250 images)} &\multicolumn{6}{c|}{CHAMELEON (76 images)} &\multicolumn{6}{c|}{COD10K (2,026 images)}&\multicolumn{6}{c}{NC4K (4,121 images)}\\ 


~ & $S_\alpha \uparrow$ & $M \downarrow$  &$E_\phi \uparrow$  & $F_\phi \uparrow $ &$F_\beta^{max} \uparrow$ &$F_\beta^w \uparrow$
& $S_\alpha \uparrow$ & $M \downarrow$  &$E_\phi \uparrow$  & $F_\phi \uparrow $ &$F_\beta^{max} \uparrow$ &$F_\beta^w \uparrow$
& $S_\alpha \uparrow$ & $M \downarrow$  &$E_\phi \uparrow$  & $F_\phi \uparrow $ &$F_\beta^{max} \uparrow$ &$F_\beta^w \uparrow$
& $S_\alpha \uparrow$ & $M \downarrow$  &$E_\phi \uparrow$  & $F_\phi \uparrow $ &$F_\beta^{max} \uparrow$ &$F_\beta^w \uparrow$
\\ \midrule

Point only & 0.6997 & 0.1012 & 0.7692 & 0.7107 & 0.7252 & 0.5880 & 0.7441 & 0.0659 & 0.7793 & 0.7180 & 0.7115 & 0.6242 & 0.7822 & 0.0399 & 0.8645 & 0.7429 & 0.7433 & 0.6679 & 0.7852 & 0.0649 & 0.8484 & 0.7947 & 0.7968 & 0.7034 \\

Text only & 0.8060 & 0.0862 & 0.8574 & 0.7682 & 0.7923 & 0.7362 & 0.8332 & 0.0590 & 0.8856 & 0.7660 & 0.7944 & 0.7510 & 0.8263 & 0.0412 & 0.8909 & 0.7327 & 0.7719 & 0.7161 & 0.8532 & 0.0518 & 0.9044 & 0.8099 & 0.8327 & 0.7867 \\ 

Segment & \textbf{0.8237} & \textbf{0.0640} & \textbf{0.8891} & \textbf{0.8032} & \textbf{0.8098} & \textbf{0.7615} & 0.8497 & 0.0403 & 0.9019 & 0.7978 & 0.8119 & 0.7699 & 0.8362 & 0.0322 & 0.9038 & 0.7541 & 0.7846 & 0.7338 & \textbf{0.8662} & \textbf{0.0399} & \textbf{0.9201} & 0.8356 & 0.8500 & 0.8086 \\ 

Segment + Choose & 0.8006 & 0.0722 & 0.8714 & 0.7877 & 0.7935 & 0.7324 & \textbf{0.8539} & \textbf{0.0395} & \textbf{0.9095} & \textbf{0.8103} & \textbf{0.8303} & \textbf{0.7835} & \textbf{0.8398} & \textbf{0.0312} & \textbf{0.9060} & \textbf{0.7624} & \textbf{0.7922} & \textbf{0.7408} & 0.8638 & 0.0411 & 0.9176 & \textbf{0.8417} & \textbf{0.8537} & \textbf{0.8088}

\\\bottomrule
\end{tabular}
}
\end{table*}

\section{Conclusion and Limitation}
In this paper, we propose a novel holistically point-guided text framework that utilizes two complementary and interactive supervision paths for WSCOD. To fully explore the potential of this dual-supervision segmentation design, novel technologies and foundation models such as Grounding DINO, SAM, CLIP, and visual engineering prompting are presented to make these two interactive and promote each other. Moreover, a bounding box rectifier and mask erasure are proposed to check and correct the mistakes for the text path supervised by point. Besides that, we design an efficient method to well choose the final pseudo label for training, which further boosts the performance. We build a new point-supervised (P2C-COD) and text-supervised (T-COD) dataset to validate the effectiveness of our proposed method. Experimental results demonstrate the superiority of our framework over the state-of-the-art weakly-supervised or unsupervised methods and narrows the gap with fully-supervised methods. Although achieving better performance compared with other SOTA methods, our method still divides the whole framework into three phases, i.e., \textit{segment, choose, train}. We leave it as future work to build an end-to-end solution.


\bibliographystyle{plain}
\bibliography{sn-bibliography}

\end{document}